\documentclass[accepted]{uai2026} 
                        
\usepackage[american]{babel}

\usepackage{natbib} 
\usepackage{mathtools} 
\usepackage{booktabs} 
\usepackage{tikz} 
\usepackage{amsmath} 
\usepackage{amssymb}
\usepackage{amsfonts}
\usepackage{amsthm} 
\usepackage{cleveref}
\usepackage{subcaption} 
\usepackage{nicefrac} 
\usepackage{wrapfig} 

\usepackage{bm}
\usepackage[ruled,vlined,linesnumbered]{algorithm2e}

\newcommand{\Rthr}{\hat{R}_{\mathrm{thr}}}
\newcommand{\Rhat}{\hat{R}}
\newcommand{\Lc}{\hat{\mathcal{L}}_C}
\newcommand{\Li}{\hat{\mathcal{L}}_I}
\newcommand{\Lp}{\hat{\mathcal{L}}_P}
\SetKwProg{Proc}{Procedure}{}{}

\newtheorem{proposition}{Proposition}[section]
\newtheorem{corollary}{Corollary}[section] 
\newtheorem{claim}{Claim}[section]

\theoremstyle{definition}
\newtheorem{definition}{Definition}[section]

\title{Particle GFlowNets: Rethinking Generative Marginalization Models}

\author[1]{\href{mailto:<tiago.dasilva@mbzuai.ac.ae.edu>}{Tiago da Silva}{}}
\author[2]{Diego Mesquita}
\author[1]{Salem Lahlou}
\affil[1]{%
    MBZUAI 
}
\affil[2]{%
    School of Applied Mathematics, Getulio Vargas Foundation
}
  
\begin{document}
\maketitle

\begin{abstract}
    Generative Marginalization Models (MaMs) have been recently introduced as efficient neural sampling models for any-order autoregressive modelling of discrete distributions. 
    By learning both the marginal and conditional probabilities of a persistent-block Gibbs sampler, MaMs enable fast posterior evaluation with a single neural network forward pass. 
    While prior work has considered MaMs to be distinct from Generative Flow Networks (GFlowNets), a well-established paradigm for inference in discrete stochastic models, we show that they are equivalent.  
    Then, we also extend MaMs' sampling strategy to non-autoregressive generative processes. 
    In particular, we describe an automatic criterion for full-state rejuvenation of the Gibbs sampler, derived from the Gelman-Rubin statistic, which plays a key role in speeding up learning convergence. 
    Our experiments 
    show that 
    our method, called Particle GFlowNets, 
    markedly accelerates 
    training 
    in large combinatorial spaces. \looseness=-1 
    
\end{abstract}

\section{Introduction} 


Generative Flow Networks \citep[GFlowNets;][]{bengio2021, foundations} have been used in recent years as flexible samplers for distributions over compositional, discrete objects such as sequences and sets. 
Notably, they were successfully applied to problems in computational biology \citep{sequence, jain2023gflownets, laajil2025curriculumaugmentedgflownetsmrnasequence}, combinatorial optimization \citep{robust, zhang2023, zhang2025hybridbalance}, and Bayesian inference \citep{deleu2022bayesian, deleu2023joint}, several of which seem not to be as easily solvable with other approaches.    
In fact, given a positive function $R(x)$ defined over a finite and combinatorial space, a GFlowNet can often generate independent samples from the corresponding probability distribution $\pi(x) \propto R(x)$ 
with remarkable accuracy \citep{shen23gflownets}. \looseness=-1 


To achieve this, GFlowNets learn a policy function for a Markov Decision Process (MDP) whose marginal distribution over terminal states matches $\pi(x)$ \citep{tiapkin2024generative}.
However, training is frequently bottlenecked by the simulation of long, overlapping trajectories, requiring countless neural network forward passes for each gradient step. 
Generative Marginalization Models \citep[MaMs;][]{liu2023mam} mitigate this issue by learning both the marginal and conditional distributions of a Gibbs sampler \citep{geman1984}.
In doing so, each learning step requires a constant number of model evaluations---corresponding to a single Gibbs transition---regardless of the state space's size. 
\looseness=-1

Contrarily to conventional understanding, we demonstrate that MaMs are equivalent to \emph{permutation-conditioned} (PC) GFlowNets, which we introduce as a particular case of the well-established conditional GFlowNet framework \citep{foundations, mogfn}.  
Specifically, we show that MaMs' learning objective corresponds to the traditional detailed balance loss function \citep[Example 5]{foundations} for PC-GFlowNets. 
With this in mind, the key contribution of MaMs is a computationally efficient Gibbs sampling algorithm that minimizes the number of neural network evaluations during training.
Their core assumption, however, is that $R(x)$ is supported on a factorized domain (i.e., $\{1, \dots, K\}^{d}$), which limits MaMs' applicability in 
permutation-invariant (e.g., graph-structured) spaces \cite[Section 3]{liu2023mam}. 
The central question 
we ask in this work is whether such an approach can be adapted to non-factorized domains. 
We answer it affirmatively. 
\looseness=-1 

For this, we propose \emph{Particle (P) GFlowNets}. 
As in MaMs, a P-GFlowNet maintains a persistent Gibbs sampler over terminal states. 
Similarly to PC-GFlowNets, a P-GFlowNet optimizes a learning objective that enforces the detailed balance condition of the underlying Markov chain. 
In contrast to both methods, which were designed for autoregressive modelling, the reverse (backward) move for P-GFlowNets is non-deterministic, as the forward mapping is non-injective. 
To circumvent this, we use a stochastic reverse transition kernel, which can also be learned.
Importantly, once trained, a P-GFlowNet can generate both independent or correlated samples via MDP simulation or Gibbs sampling, respectively.
The crucial difference lies in the per-generation computational cost and the effective sample size of the generated samples, both of which are larger for independent sampling.  
Given a limited time budget, the optimal approach will likely need to be determined in a case-by-case basis. 
\looseness=-1


In practice, we find that samples from P-GFlowNet's persistent Gibbs chain can become highly correlated during training, reducing the accuracy of gradient estimates and slowing down convergence.
To address this limitation, we occasionally refresh the stochastic process based on the Gelman-Rubin statistics \cite{GelmanRubin1992, carpenter2017stan}, also known as R-hat or $\hat{R}$, being above a certain threshold. 
We empirically show that this process, called \emph{rejuvenation} in the probabilistic programming literature \cite{lew2022pcleanbayesiandatacleaning, lew2022recursivemontecarlovariational}, significantly accelerates learning convergence.  
\looseness=-1 


Additionally, we evaluate P-GFlowNets on both established benchmarks 
and 
novel tasks for which relevant functionals can be efficiently computed and used for reliable performance assessment. 
Our experiments show that P-GFlowNets significantly accelerate training convergence when the cost of MDP simulation exceeds that of evaluating the target distribution, which is typical for large, long-horizon state spaces. 
In summary, our contributions are as follows. 

\looseness=-1 
\begin{enumerate}[leftmargin=12pt]
    \item We show MaMs can be 
    represented as PC GFlowNets, and that their learning objective is equivalent to \citet{foundations}'s expected detailed balance loss function.
    \item We introduce P-GFlowNets. 
    When compared 
    to prior methods 
    for GFlowNet training,
    P-GFlowNets asymptotically reduce the average number of 
    forward passes per gradient step as the MDP's horizon increases. \looseness=-1 
    \item We empirically demonstrate that P-GFlowNets 
    significantly reduce wall-clock time for learning convergence. \looseness=-1 
\end{enumerate}

The reader only interested in learning about P-GFlowNets may skip \Cref{sec:equivalenceas} and focus on \Cref{sec:pgflownets,sec:experiments,sec:summaryas} instead.  
\looseness=-1 



\section{Preliminaries} \label{sec:background} 

\paragraph{Notations.} 
Let $\mathcal{X}$ be a discrete space.
We assume $\mathcal{X}$ is \emph{compositional}, i.e., each $x \in \mathcal{X}$ can be described as a collection of \emph{components} $\mathcal{C}$, $x = \{c_{1}, \dots, c_{d}\} \subseteq \mathcal{C}$, with $d \ge 1$ possibly depending on $x$. 
Often, we will consider $\mathcal{X} \coloneqq [K]^{d} \coloneqq \{1, \dots, K\}^{d}$ for positive integers $K$ and $d$, in which case we will refer to $d$ as $\mathcal{X}$'s dimension.
In this setting, $\mathcal{C} = [K] \times [d]$ and each $x = \{(i_{1}, 1), \dots, (i_{d}, d)\}$ corresponds to the sequence $(i_{1}, \dots, i_{d}) \in [K]^{d}$. 
We use these notations interchangeably, 
i.e., $x = \{(1, 1), (0, 2), (1, 3)\}$ represents the sequence $(1, 0, 1)$, with $x_{1} = 1$, $x_{2} = 0$, and $x_{3} = 1$. 
In particular, we write $x \equiv \{(i_{j}, j)\}_{j=1}^{d}$.
In general, $\mathcal{X} \subseteq 2^{\mathcal{C}}$ is a subset of $\mathcal{C}$'s power set, $2^{\mathcal{C}}$. 
\looseness=-1 

\begin{figure}[!h]
    \centering
    \begin{subfigure}{.35\linewidth} 
        \centering 
        \includegraphics[width=.9\linewidth, page=3]{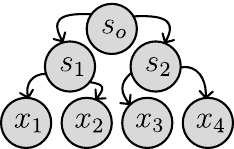}
        \caption{Autoregressive.}
    \end{subfigure}
    \hspace{18pt}
    \begin{subfigure}{.4\linewidth} 
        \centering 
        \includegraphics[width=.9\linewidth, page=4]{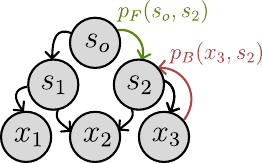} 
        \caption{Non-autoregressive.}
    \end{subfigure}
    \caption{The state graph ($\mathcal{G}$) is a tree in an autoregressive setting (a). When many trajectories lead to the same state, such as $s_{o} \rightarrow (s_{1}, s_{2}) \rightarrow x_{2}$ in (b), $\mathcal{G}$ is a non-tree DAG. \looseness=-1 }
    \label{fig:stategraphs}
    \vspace{-12pt} 
\end{figure}

Our objectives are to generate $x \in \mathcal{X}$ in proportion to a given positive measure in $\mathcal{X}$, the probability mass of which we will denote by $R \colon \mathcal{X} \rightarrow \mathbb{R}_{+}$, and to evaluate the marginal probability of subsets of $x$ under $R$. 
We will call $\pi(x) \propto R(x)$ the normalized distribution. 
As in \cite{bengio2021, malkin2023gflownets}, we refer to $R$ as a \emph{reward function}.
Additionally, we will define the space $\mathcal{S}$ as 
\begin{equation} \label{eq:projections} 
    \mathcal{S} = \{s \in 2^{\mathcal{C}} \colon s \subset x \text{ for some } x \in \mathcal{X}\}. 
\end{equation}
Clearly, $\emptyset \in \mathcal{S}$.
As in \cite{bengio2021,foundations}, we call $\mathcal{S} \cup \mathcal{X}$ the \emph{state space}, and define $s_{o} \coloneqq \emptyset$ as the \emph{initial state}.
Similarly, we let $\mathcal{G} \coloneqq (\mathcal{S} \cup \mathcal{X}, \mathcal{E})$, $\mathcal{E} = \{(s, s') \colon \exists c \in \mathcal{C} \setminus s' \text{ such that } s' = s \cup \{c\}\}$, be the \emph{state graph}. 
Intuitively, $s \rightarrow s'$ in $\mathcal{G}$ if $s'$ differs from $s$ by a single additional component. 
We say that the generative process is \emph{autoregressive} when $\mathcal{G}$ is a tree rooted at $s_{o}$; see \Cref{fig:stategraphs}. \looseness=-1



\paragraph{GFlowNets.} 
A GFlowNet learns a \emph{forward} policy function $p_{F} \colon \mathcal{S} \times (\mathcal{S} \cup \mathcal{X}) \rightarrow [0, 1]$ such that $p_{F}(s, \cdot)$ is a probability distribution supported on the children of $s$ in the state graph $\mathcal{G}$.
By defining $s_{o} \rightsquigarrow x$ as the set of trajectories in $\mathcal{G}$ from $s_{o}$ to $x \in \mathcal{X}$, we seek a function $p_{F}$ satisfying \looseness=-1    
\begin{equation} \label{eq:marginal} 
    p_{\top}(x) \coloneqq \sum_{\tau \in s_{o} \rightsquigarrow x} p_{F}(s_{o}, \tau) \propto R(x), 
\end{equation}
with $\tau = (s_{o}, \dots, s_{d - 1}, x)$, $s_{d} \coloneqq x$, and (with a slight abuse of notation) $p_{F}(s_{o}, \tau) = \prod_{i=1}^{d} p_{F}(s_{i - 1}, s_{i})$; we refer to $d$ as the trajectory's length.
\Cref{eq:marginal} represents the marginal distribution over $\mathcal{X}$ induced by $p_{F}(s_{o}, \cdot)$. 
When the generative process is autoregressive, $s_{o} \rightsquigarrow x$ contains a single trajectory $\tau_{x}$, and $p_{\top}(x) = p_{F}(s_{o}, \tau_{x})$.
Otherwise, the set $s_{o} \rightsquigarrow x$ might be intractably large; see \Cref{fig:stategraphs}. 
Under these conditions, we introduce a \emph{backward policy} $p_{B}$, which is a forward policy on the transposed state graph, and approximate \Cref{eq:marginal} via importance sampling, 
\begin{equation*}
    p_{\top}(x) = \mathbb{E}_{\overleftarrow{\tau} \sim p_{B}(x, \cdot)} \left[ \frac{p_{F}(s_{o}, \overrightarrow{\tau})}{p_{B}(x, \overleftarrow{\tau})} \right] \propto R(x), 
\end{equation*}
in which we use $\overrightarrow{\tau}$ and $\overleftarrow{\tau}$ to distinguish the forward and backward trajectories.
(We will drop this notation in the rest of the text for clarity). 
By letting $Z \coloneqq \sum_{x \in \mathcal{X}} R(x)$ be the partition function of our target distribution, the above equation may be re-arranged as (dividing both sides by $R(x)$)
\begin{equation} \label{eq:tb} 
    \mathbb{E}_{\tau \sim p_{B}(x, \cdot)}\left[ \frac{Z \cdot p_{F}(s_{o}, \tau)}{R(x) \cdot p_{B}(x, \tau)} \right] = 1. 
\end{equation}
The condition $Z \cdot p_{F}(s_{o}, \tau) = R(x) \cdot p_{B}(x, \tau)$ is known as \emph{trajectory balance} \citep[TB;][]{malkin2022trajectory}. 
In practice, $p_{F}(s, \cdot)$ is parameterized as a softmax neural network receiving $s$ as input, and we optimize both $p_{F}$ and $Z$ via stochastic gradient descent on the objective function 
\begin{equation*}
   \mathcal{L}_{\mathrm{TB}}(p_{F}, p_{B}, Z) = \mathbb{E}_{\tau \sim p_{E}} \left[ \left( \log \frac{Z \cdot p_{F}(s_{o}, \tau)}{R(x) \cdot p_{B}(x, \tau)} \right)^{2} \right],  
\end{equation*}
with $p_{E}$ as an exploratory (full-support) policy \citep[see, e.g.,][]{madan2025towards, kim2025adaptive}. As the reader may have noticed, $p_{B}$ was chosen arbitrarily; it can be either learned \citep[e.g.,][]{shen23gflownets, gritsaev2025optimizingbackwardpoliciesgflownets} or fixed 
\citep[e.g.,][]{deleu2022bayesian, zhou2024phylogfn}.
\looseness=-1 

Alternatively, we commonly parameterize the \emph{flow functions} 
\begin{equation*}
    F(s) = Z \sum_{\tau \in s_{o} \rightsquigarrow s} p_{F}(s_{o}, \tau),
\end{equation*}
for $s \neq s_{o}$ and $F(s_{o}) = Z$, and define the \emph{detailed balance} (DB) learning objective 
$\mathcal{L}_{\mathrm{DB}}(p_{F}, p_{B}, F)$ as  
\begin{equation} \label{eq:db} 
    \mathbb{E}_{\tau \sim p_{E}} \left[ \sum_{1 \le i \le d} w_{i} \left ( \log \frac{F(s_{i - 1}) p_{F}(s_{i - 1}, s_{i})}{F(s_{i}) p_{B}(s_{i}, s_{i - 1})} \right)^{2} \right], 
\end{equation}
with $\tau = (s_{o}, \dots, s_{d})$ as in \Cref{eq:marginal}, $F(s_{d}) = R(s_{d})$, and $w_{i} > 0$ as positive weights such that $\sum_{i=1}^{n} w_{i} = 1$. 
Common choices for $w_{i}$ are either uniform $w_{i} = \nicefrac{1}{n}$ \citep{foundations} or $w_{i} \propto \exp\{\lambda i\}$ for $\lambda > 0$, the intuition being that states closer to $\mathcal{X}$ should be assigned with larger weights during training \citep{silva2025when}.  
Other learning objectives have also been studied \citep[e.g.,][]{madan2022learninggf, robust}; see \Cref{sec:relatedworks} for related works. 

A \emph{conditional} GFlowNet, on the other hand, models a family of distributions $R_{\omega}(x)$ indexed by a parameter $\omega \in \Omega$ that is also used as input for both the policy ($p_{F}$ and $p_{B}$) and flow ($F$) functions. 
Prominent applications include multi-objective combinatorial optimization, in which $\Omega$ is a simplex and $R_{\omega}$ represents a weighted mixture of each objective according to $\omega$ \citep{mogfn, roy2023goal, zhu2023sampleefficient, laajil2025curriculumaugmentedgflownetsmrnasequence}, and distributed learning, with each $\omega \in \Omega$ representing a distinct subgraph of the state graph and $R_{\omega}(x)$ simply referring to the restriction of a given $R(x)$ to the corresponding subset of $\mathcal{X}$ \citep{silva2025generalization}.
\looseness=-1 

\paragraph{MaMs.} 
In the context of sampling, MaMs were designed by \cite{liu2023mam} for fast autoregressive modelling of discrete models.
Simply put, they 
concomitantly learn a marginal and conditional distributions, $p_{\theta}$ and $p_{\phi}$, on $[K]^{d}$ satisfying both a \emph{correctness} and \emph{consistency} conditions. 
Given indices $\mathcal{I}, \mathcal{J} \subseteq \{1, \dots, d\}$ such that $\mathcal{I} \subset \mathcal{J}$ and $|\mathcal{J} \setminus \mathcal{I}| = M$, 
we respectively define these conditions as \looseness=-1  
\begin{equation} \label{eq:mams} 
    p_{\theta}(x) = \frac{R(x)}{Z} \text{ and } p_{\theta}(x_{\mathcal{I}}) p_{\phi}(x_{\mathcal{J}}|x_{\mathcal{I}}) = p_{\theta}(x_{\mathcal{J}}), 
\end{equation}
for $x \in \mathcal{X}$. As in \cite{liu2023mam}, we assume $M = 1$, although our analysis can be easily extended to $M > 1$ by interpreting a group of $M$ adjacent variables as a single variable.
As with GFlowNets, $p_{\theta}$ and $p_{\phi}$ are parameterized as neural networks.
Also, a special symbol $\triangle$ is used to represent a marginalized variable---e.g., if $x = (x_{1}, \dots, x_{d})$, then $x_{\{1, 2\}} = (x_{1}, x_{2}, \triangle, \dots, \triangle)$---and $Z$ is a learned parameter.  
Once trained, the probability $p_{\theta}(x_{\mathcal{I}})$ of any $x \in \mathcal{X}$ and any $\mathcal{I} \subseteq \{1, \dots, d\}$ can be evaluated in a single neural network forward pass.
To enforce \Cref{eq:mams}, we minimize 
\begin{equation} \label{eq:mamsobjectives} 
    \mathrm{KL}[p_{\theta} || \pi] + \lambda \mathrm{ConsistencyError}(p_{\theta}, p_{\phi}, Z), 
\end{equation}
in which $\lambda > 0$, $\mathrm{KL}[p_{\theta}||\pi] \coloneqq \mathbb{E}_{x \sim p_{\theta}}\left[ \log \frac{p_{\theta}(x)}{\pi(x)} \right]$ is the Kullback-Leibler (KL) divergence between $p_{\theta}$ and $\pi$, and $\mathrm{ConsistencyError}$ is defined as \looseness=-1  
\begin{equation*}
    \mathbb{E}_{x \sim q} \mathbb{E}_{m} \mathbb{E}_{\sigma} \left( \log \frac{p_{\theta}(x_{\sigma( < m)}) \cdot p_{\phi}(x_{m}|x_{\sigma([m - 1])})}{p_{\theta}(x_{\sigma([m])})}  \right)^{2}, 
\end{equation*}
with $m \sim \mathcal{U}[d]$, $\sigma \sim \mathcal{U}(S_{d})$, $q$ as any distribution over $\mathcal{X}$, $\mathcal{U}[d]$ as an uniform distribution over $[d] = \{1, \dots, d\}$, (hard-coded) restriction $p_{\theta}((\triangle, \dots, \triangle)) = Z$, and $\sigma$ uniformly picked from the space of permutations, which we denote by 
\begin{equation} \label{eq:permutations} 
    S_{d} = \{ \sigma \colon [d] \rightarrow [d] \ | \ \sigma \text{ is bijective} \}. 
\end{equation} 
To generate samples from $p_{\theta}$, which are needed to estimate the KL divergence in \Cref{eq:mamsobjectives}, \cite{liu2023mam} implement a persistent-block Gibbs sampling scheme having $p_{\phi}$ as the transition kernel. 
As we explain next, under the consistency condition in \Cref{eq:mams}, the resulting Markov chain has the marginal $p_{\theta}$ as the stationary distribution. 


\paragraph{Gibbs sampling.} 
Originally designed by \citep{geman1984} to sample from the Gibbs distribution \citep{gibbs1902elementary}, and also known as Successive Substitution Sampling \citep[SSS;][]{schervish2012theory}, Gibbs sampling creates a Markov chain by iteratively modifying each coordinate of a state $x$ according to the conditionals of the target distribution $\pi(x)$. 
That is, we iteratively select $i \in [d]$ and then $x_{i}' \sim \pi(x_{i} | x_{-i})$, with $x_{-i} = x_{\{1, \dots, i - 1, i + 1, \dots, d\}}$ as in \Cref{eq:mams}. 
This process may be pictured as $(x_{1}, x_{2}) \rightarrow (x_{1}', x_{2}) \rightarrow (x_{1}', x_{2}')$ for two-dimensional distributions. 
When $i$ is picked at random, the algorithm is known as \emph{random scan} Gibbs sampler; otherwise, as \emph{systematic scan} Gibbs sampler \citep{mcbook}. 
We will focus on the former. 
The reader is invited to notice that the resulting Markov chain is stationary with respect to the target distribution $\pi$. \looseness=-1 

In the context of learning and energy-based models \citep{hinton2002, tieleman2008training, liu2023mam}, a \emph{persistent-block} Gibbs sampler extends the canonical algorithm by (i) grouping variables into blocks, each of which is updated in a single step, and (ii) maintaining the Markov chain's state across gradient updates of the underlying neural sampling model.
We notice that, in this case, the Markov chain ceases to be stationary until training converges.
\looseness=-1

\section{Rethinking MaMs as GFlowNets}
\label{sec:equivalenceas}

We now show that MaMs may be interpreted as a conditional GFlowNet (\Cref{sec:pcgflownets}).
In a nutshell, the conditioning space $\Omega \coloneqq S_{d}$ will be the space of permutations defined in \Cref{eq:permutations}, and we will demonstrate the consistency loss function is equivalent to the weighted DB loss in \Cref{eq:db} in expectation (\Cref{sec:mamsgflownets}). \looseness=-1 

\subsection{Permutation-conditioned GFlowNets} \label{sec:pcgflownets}  

\begin{figure}[!h]
    \centering
    \begin{subfigure}{.4\linewidth}  
        \includegraphics[width=\linewidth, page=1]{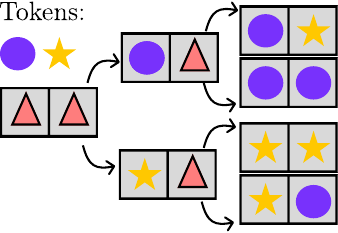}
        \caption{$\sigma(1) = 1, \sigma(2) = 2$.}
    \end{subfigure}
    \hspace{18pt} 
    \begin{subfigure}{.4\linewidth}
        \includegraphics[width=\linewidth, page=2]{uai2026-template/figures/stategraphs.pdf}
        \caption{$\sigma(1) = 2, \sigma(2) = 1$.}
    \end{subfigure}
    \caption{
    \textbf{A PC GFlowNet fills up a sequence according to a given permutation} ($\sigma$). After trained, it supports fast evaluation of any-subset marginals. 
    ($\triangle$ means no token.)
    \looseness=-1}
    \label{fig:pcgflownets}
\end{figure}

To understand the connection between MaMs and GFlowNets, we first introduce a family of \emph{permutation-conditioned} (PC) GFlowNets for autoregressive modelling.
Again, let $\mathcal{X} = [K]^{d}$.
We define a learnable \emph{state-embedding function} $\psi \colon \mathcal{S} \rightarrow \mathbb{R}^{h}$ and weights $\{\mathbf{W}_{k}\}_{k=1}^{K}$ such that $\mathbf{W}_{k} \in \mathbb{R}^{h \times d}$ for a given dimension $h$. 
Then, recall from \Cref{sec:background} that the state space $\mathcal{S} \cup \mathcal{X}$ is simply a subset of the power set of $\mathcal{C} \coloneqq [K] \times [d]$. 
As such, the policy function evaluated at $s^{(i)} \coloneqq \{(k_{1}, \sigma(1)), \dots, (k_{i}, \sigma(i))\} \in \mathcal{S}$ and conditioned on a permutation $\sigma \in S_{d}$ is defined as  
\begin{equation} \label{eq:policies} 
    p_{F}^{\sigma}(s^{(i)}, s^{(i + 1)}) \propto  \exp\{ \mathbf{w}_{k, \sigma(i + 1)}^{T} \psi(s^{(i)}) \}
\end{equation}
for $s^{(i + 1)} = s^{(i)} \cup \{ (k, \sigma(i + 1) )\}$ and $k \in [K]$.  Intuitively, $p_{F}^{\sigma}$ fills up a sequence up to a prescribed size ($d$, in this case) according to the ordering imposed by $\sigma$. 
We illustrate this in 
\Cref{fig:pcgflownets}.
Given a $\sigma$ and a $s^{(i + 1)} \in \mathcal{S}$, there is only one state $s^{(i)}$ such that the transition $s^{(i)} \rightarrow s^{(i + 1)}$ has positive probability under $p_{F}^{\sigma}(s_{i}, \cdot)$. 
Hence, the only choice for the backward policy abiding by the TB condition in \Cref{eq:tb} is $p_{B}^{\sigma}(s^{(i + 1)} , \cdot) = \delta_{s^{(i)} }$ for $s^{(i)} = \{(k_{j}, \sigma(j))\}_{j=1}^{i}$ and $s^{(i + 1)} = s^{(i)} \cup \{(k_{i + 1}, \sigma(i + 1))\}$ with $(k_{j})_{j=1}^{i + 1} \in [K]^{i + 1}$; $\delta_{s^{(i)}}$ is the Dirac delta at $s^{(i)}$, i.e., $\delta_{s^{(i)}}(s) = 1$ if $s = s^{(i)}$ and $0$ otherwise.
Thus, as $p_{B}$ plays no significant role in learning, we set it aside from most of our analysis until \Cref{sec:pgflownets}.
Our main claim in this section is that there a bijective correspondence between MaMs and PC GFlowNets.
\looseness=-1 

\begin{claim} \label{claim:mams} 
    For each MaM, there is a unique and functionally equivalent PC GFlowNet; see \Cref{prop:equivalences}.   
\end{claim}

Clearly, the conditioning permutation $\sigma$ only affects the navigation of $\mathcal{S}$, as both $\mathcal{X}$ and $R \colon \mathcal{X} \rightarrow \mathbb{R}_{+}$ remain unchanged under $\sigma$ (except for re-labeling). 
Interestingly, however, the permutation-conditioned flow function $F^{\sigma}$ satisfying the DB condition in \Cref{eq:db} can be exactly interpreted as the marginal probability distribution of a sequence under $R$. 
This result, outlined below, is crucial in bridging the gap between GFlowNets and MaMs. \looseness=-1 

\begin{proposition} \label{prop:marginals} 
    Let $F^{\sigma}$ be a flow function abiding by the DB condition of a permutation-conditioned GFlowNet, i.e., 
    \begin{equation} \label{eq:dbsigma} 
        F^{\sigma}(s) p_{F}^{\sigma}(s, s') = F^{\sigma}(s') p_{B}^{\sigma}(s', s)
    \end{equation}
    and $F^{\sigma}(x) = R(x)$ for $x \in \mathcal{X}$.
    Then, denoting by $s^{(i)} = \{(k_{1}, \sigma(1)), \dots, (k_{i}, \sigma(i))\}$ for $i \le d$, 
    \begin{equation} \label{eq:dbmarginals}
        F^{\sigma}(s^{(i)}) =  \!\!\!\!\!\! \sum_{x \in \mathcal{X} \colon x_{\sigma([i])} = s^{(i)}_{\sigma([i])}} \!\!\!\!\!\! R(x) \text{ and } F^{\sigma}(s_{o}) = \sum_{x \in \mathcal{X}} R(x), 
    \end{equation}
    in which $x_{\sigma([i])} = s^{(i)}_{\sigma([i])}$ means that $x \in [K]^{d}$ agrees with $s^{(i)}$ on the first $i$ coordinates according to the permutation $\sigma$, i.e., $x_{\sigma(1)} = s_{\sigma(1)}^{(i)}, \dots, x_{\sigma(i)} = s_{\sigma(i)}^{(i)}$. 
    In other words, $\frac{F^{\sigma}(s^{(i)})}{F^{\sigma}(s_{o})}$ is the marginal distribution of $s^{(i)}$ under $R$. \looseness=-1 
\end{proposition}

As in \Cref{sec:background}, the state $s^{(i)}$ in the \Cref{prop:marginals} can be written as $s_{j}^{(i)} = k_{j}$ for $j \in \sigma([i])$ and $s_{j}^{(i)} = \triangle$ otherwise. \looseness=-1 

When learning a conditional GFlowNet, we often augment the input of the neural networks parameterizing both $F^{\sigma}$ and $p_{F}^{\sigma}$ with $\sigma$ \citep{foundations, robust}.
As MaMs avoid such augmentation, we ask: is it needed for PC GFlowNets?
The following corollary shows that, when states $s^{1}$ and $s^{2}$ differ solely by the 
permutation $\sigma$, the optimal flow function in \Cref{eq:dbsigma} does not depend on $\sigma$. \looseness=-1 

\begin{corollary} \label{col:marginals} 
    Let $\sigma_{1}, \sigma_{2} \in S_{d}$ be permutations, and define $s^{1, (i)} = \{(k_{1}, \sigma_{1}(1)), \dots, (k_{i}, \sigma_{1}(i))\}$ and $s^{2, (i)} = \{(k_{1}, \sigma_{2}(1)), \dots, (k_{i}, \sigma_{2}(i))\}$. 
    Then, if $i$ satisfies $\sigma_{1}([i]) = \sigma_{2}([i])$, i.e., $\sigma_{1}$ and $\sigma_{2}$ coincide on $[i] \coloneqq \{1, \dots , i\}$, \looseness=-1 
    \begin{equation*}
        F^{\sigma_{1}}(s^{1, (i)}) = F^{\sigma_{2}}(s^{2, (i)}). 
    \end{equation*}
\end{corollary}

To evaluate $p_{F}^{\sigma}(s^{(i)}, s^{(i + 1)})$, however, information about $\sigma(i + 1)$ is needed. 
As we show below, 
$\sigma_{1}$ and $\sigma_{2}$ coincide on $[i + 1]$, then $p_{F}^{\sigma_{1}}(s^{(i)}, s) = p_{F}^{\sigma_{2}}(s^{(i)}, s)$ for each $s \in \mathcal{S} \cup \mathcal{X}$.  

\begin{corollary} \label{col:conditionals} 
    Let $\sigma_{1}, \sigma_{2} \in S_{d}$. 
    Define $s^{1, (i)}$ and $s^{2, (i)}$ as in \Cref{col:marginals}. 
    Assume $\sigma_{1}(\le i + 1) = \sigma_{2}(\le i + 1)$. 
    Then,
    \looseness=-1 
    \begin{equation*}
        p_{F}^{\sigma_{1}}(s^{1, (i)}, \cdot) = p_{F}^{\sigma_{2}}(s^{2, (i)}, \cdot), 
    \end{equation*}
    i.e., if $s^{1, (i + 1)} \equiv s^{2, (i + 1)}$ correspond to the same object in $([K] \cup \{\triangle\})^{d}$, $p_{F}^{\sigma_{1}}(s^{1, (i)}, s^{1, (i + 1)}) = p_{F}^{\sigma_{2}}(s^{2, (i)}, s^{2, (i + 1)})$. 
\end{corollary}

In practice, we parameterize $p_{F}^{\sigma}(s^{(i)}, \cdot)$ as a neural network returning a matrix $\mathbf{P} = \mathbb{R}_{+}^{K \times d}$ with $\mathrm{sum}(\mathbf{P}) = 1$ representing the probability of each variable $k$ being in the $i$th position, $\mathbf{P}_{k, i}$, and mask out all columns except the $\sigma(i + 1)$-th one, which results in \Cref{eq:policies}.
Drawing on the derivations above, \Cref{claim:mams} is only a matter of bookkeeping. 
\looseness=-1 






\paragraph{Notation mapping.} 
We recall 
a MaM is characterized by a marginal $p_{\theta}$ and conditional $p_{\phi}$ distributions. 
Also, for each 
$s = (k_{1}, \dots, k_{d})$ with $k_{j} \in [K]$ or $k_{j} = \triangle$ (i.e., $k_{j}$ is marginalized), as described in \Cref{sec:background}, there is a permutation $\sigma$ and an index $i$ for which $s \equiv \{(j, \sigma(j))\}_{j=1}^{i}$ in the PC GFlowNet's state graph. 
By \Cref{col:marginals}, we can unambiguously define $F^{\sigma}(s) \coloneqq p_{\theta}(s)$. 
Similarly, let $\mathcal{J} \supset \mathcal{I}$ be subsets of $\{1, \dots, d\}$, and 
define $I = |\mathcal{I}|$. (Recall $|\mathcal{J} \setminus \mathcal{I}| = 1$). 
As before, given $x \in \mathcal{X}$, there is a permutation $\sigma$ such that $\sigma([I]) = \mathcal{I}, \sigma([I + 1]) = \mathcal{J}$, and \looseness=-1 
\begin{equation} \label{eq:sequencesstates} 
    \begin{aligned} 
        x_{\mathcal{I}} \equiv \{(x_{\sigma(j)}, \sigma(j))\}_{j=1}^{I} \text{ and } x_{\mathcal{J}} \equiv \{(x_{\sigma(j)}, \sigma(j))\}_{j=1}^{I + 1}. 
    \end{aligned} 
\end{equation}
Under \Cref{col:conditionals}, $p_{F}^{\sigma}(x_{\mathcal{I}}, \cdot)$ does not depend on $\sigma(j)$ for $j > I + 1$. 
Again, we can thus define $p_{\phi}(x_{\mathcal{J}}|x_{\mathcal{I}}) = p_{F}^{\sigma}(x_{\mathcal{I}}, x_{\mathcal{J}})$ without ambiguity.
Our central result, stated below, establishes that there is a unique PC GFlowNet for each MaM satisfying both the distributional (i.e., $p_{\theta}(x) = \pi(x)$ for $x \in \mathcal{X}$) and consistency conditions, and vice-versa. 
\looseness=-1  

\begin{proposition}[MaMs are PC GFlowNets] \label{prop:equivalences} 
    Let $(p_{\theta}, p_{\phi})$ be a consistent MaM satisfying $p_{\theta}(x) = \pi(x)$.
    Then, there is a unique PC GFlowNet such that $F^{\sigma}(x_{\mathcal{J}}) = p_{\theta}(x_{\mathcal{J}})$ and $p_{F}^{\sigma}(x_{\mathcal{I}}, x_{\mathcal{J}}) = p_{\phi}(x_{\mathcal{J}} | x_{\mathcal{I}})$ for each triplet $\mathcal{I}, \mathcal{J}, \sigma$ as in \Cref{eq:sequencesstates}. 
    This PC GFlowNet 
    satisfies the DB condition in \Cref{eq:dbsigma}. 
    Conversely, a PC GFlowNet satisfying DB induces a unique consistent MaM such that $p_{\theta}(x) = \pi(x)$. \looseness=-1 
\end{proposition}

From an operational viewpoint, both MaMs and PC GFlowNets allow for the evaluation of any-subset marginals with a single neural network forward pass in discrete models. 
In fact, the only lingering difference between MaMs and PC GFlowNets is their differing learning objectives.
\looseness=-1 

As we discuss next, however, the estimator used for MaM's loss function is biased unless the consistency condition holds. 
Hence, it might be unsuited for training. 
We instead derive a novel unbiased estimator that preserves MaMs' constant number of forward passes per gradient step.  




\subsection{Revisiting MaMs' objective} \label{sec:mamsgflownets} 

At a basic level, MaMs aim to minimize the KL divergence between $p_{\theta}$ and 
$\pi$ under the consistency constraint, 
i.e., \looseness=-1 
\begin{equation*}
    \min_{p_{\theta}} \mathrm{KL}[p_{\theta} || \pi] \text{ such that } p_{\theta}(x_{\mathcal{I}}) p_{\phi}(x_{\mathcal{J}}|x_{\mathcal{I}}) = p_{\theta}(x_{\mathcal{J}}) 
\end{equation*}
for each set of indices $\mathcal{I}, \mathcal{J}$ as in \Cref{prop:equivalences}.
As directly enforcing $p_{\theta}$ to be consistent with $p_{\phi}$ is computationally unfeasible, a penalty function---the $\mathrm{ConsistencyError}$---is introduced instead; recall \Cref{eq:mamsobjectives}.
A central question, however, remains unanswered: how to generate samples from $p_{\theta}$ to estimate the objective function $\mathrm{KL}[p_{\theta} || \pi]$? 

When the consistency constraint is satisfied, a persistent Gibbs sampling scheme using $p_{\phi}$ as the transition kernel would realize a Markov chain ergodic with respect to $p_{\theta}$, the samples of which could be used for estimating $\mathrm{KL}[p_{\theta}||\pi]$.
However, $p_{\theta}$ is only approximately consistent with $p_{\phi}$.
Thus, 
the generated samples would produce biased estimates of the KL objective. 
Based on the connection between MaMs and PC GFlowNets in \Cref{prop:equivalences}, we propose instead 
minimizing the consistency-only objective below.
\looseness=-1 

\begin{definition} \label{def:consistency} 
    We let $p_{\theta}(x_{\mathcal{I}}) = \frac{F_{\theta}(x_{\mathcal{I}})}{Z}$ for 
    given parametric function $F_{\theta}$ and 
    $x \in \mathcal{X}$ and indices $\mathcal{I} \subseteq [d]$ with $Z \coloneqq F(x_{\emptyset})$ learnable. 
    The consistency-only objective $\mathcal{L}_{\mathrm{CO}}$ is \looseness=-1 
    \begin{equation*}
       \mathbb{E}_{x \sim q, m, \sigma} 
       \left( \log \frac{F_{\theta}(x_{\sigma([m - 1])}) p_{\phi}(x_{\sigma(m)}|x_{\sigma([m - 1])})}{F_{\theta}(x_{\sigma([m])})} \right)^{2},
    \end{equation*}
    in which $\sigma \sim \mathcal{U}(S_{d})$, $m \sim \mathrm{Cat}(\mathbf{w})$ for $\mathbf{w} \ge 0$ and $\sum_{i=1}^{d} \mathbf{w}_{i} = 1$, and $q$ is a full-support distribution over $\mathcal{X}$. 
    Additionally, we restrict $F_{\theta}(x) = R(x)$ for $x \in \mathcal{X}$. 
\end{definition}

The only difference between $\mathcal{L}_{\mathrm{CO}}$ and MaMs' consistency error is that we enforce $F_{\theta}(x) = R(x)$ for $x \in \mathcal{X}$ instead of minimizing $\mathrm{KL}[p_{\theta}||\pi]$. 
Clearly, when $\mathcal{L}_{\mathrm{CO}}$ is globally minimized, $F_{\theta}(x_{\emptyset}) = \sum_{x \in \mathcal{X}} R(x)$, which allows for direct evaluation of $p_{\theta}$. 
Moreover, the $\mathcal{L}_{\mathrm{CO}}$ corresponds to the DB loss $\mathcal{L}_{\mathrm{DB}}$ in \Cref{eq:db} with a specific weight scheme $(w_{i})_{i=1}^{d}$.
In contrast to $\mathcal{L}_{\mathrm{DB}}$, however, a Monte Carlo approximation of $\mathcal{L}_{\mathrm{CO}}$ requires a constant number of 
forward passes with respect to the state space's dimension. \looseness=-1 

\begin{proposition} \label{prop:dbcon} 
    Let $(F, p_{F}^{\sigma})$ be a PC GFlowNet. 
    (By \Cref{col:marginals}, $F$ does not depend on $\sigma$).
    Let $p_{E}^{\sigma}$ be an exploratory policy such that the marginal 
    of $p_{E}^{\sigma}(s_{o}, \cdot)$ over $\mathcal{X}$ matches the distribution $q$ defined in \Cref{def:consistency}. 
    Then, 
    \begin{equation*} \label{eq:dbcon}
        \begin{aligned} 
            &\underset{\substack{{\sigma \sim \mathcal{U}(S_{d})}, \\ \tau \sim p_{E}^{\sigma}(s_{o}, \cdot)}}{\mathbb{E}} \left[ w_{i} \sum_{1 \le i \le d} \left( \log \frac{F(s_{i - 1}) p_{F}^{\sigma}(s_{i} | s_{i - 1})}{F(s_{i})} \right)^{2} \right] = \\ 
            &\underset{\substack{\sigma \sim \mathcal{U}(S_{d}), \ x \sim q, \\ i \sim \mathrm{Cat}(\mathbf{w})}}{\mathbb{E}} 
            \left[ \left( \log \frac{F(x_{\sigma([i - 1])}) p_{F}^{\sigma}(x_{\sigma([i - 1])}, x_{\sigma([i])})}{F(x_{\sigma([i])})} \right)^{2} \right], 
        \end{aligned} 
    \end{equation*}
    with $\tau = (s_{i})_{i=0}^{d}$, 
    $s_{d} = x$, 
    and $x_{\sigma([i])} = \{(x_{j}, \sigma(j))\}_{j=1}^{i}$.  
\end{proposition}

\Cref{prop:dbcon} establishes that 
MaMs' $\mathrm{ConsistencyError}$ 
corresponds to a transition-wise estimator of the conventional DB loss function for conditional GFlowNets. 
After thoroughly 
outlining the connection between PC GFlowNets and MaMs, we ask: how can we leverage our results to improve GFlowNet training? 
We investigate this next. \looseness=-1 

    





\section{Particle GFlowNets} \label{sec:pgflownets} 

We extend 
the persistent-block Gibbs sampling scheme to non-autoregressive generative processes, such as set generation, in which multiple trajectories may lead to 
the same object.
As with MaMs, this reduces the number of 
forward passes from $\mathcal{O}(d)$ to $\mathcal{O}(1)$ per gradient step. 
We also develop an automatic criterion for refreshing the 
Gibbs chain, which improves exploration and accelerates convergence. 
\looseness=-1 

\paragraph{Particle GFlowNets.} 
We recall from \Cref{sec:background} that a compositional object $x$ is represented as a collection of components from a set $\mathcal{C}$, i.e., $x = \{c_{1}, \dots, c_{d}\}$. 
In this scenario, a policy function induces a probability distribution over a subset of $\mathcal{C}$ conditioned on a state $s \in 2^{\mathcal{C}}$. 
For PC GFlowNets, $\mathcal{C} = [K] \times [d]$ and $p_{F}^{\sigma}(s^{(i)}, \cdot)$ induces a distribution over $[K] \times \{\sigma(i + 1)\} \subset \mathcal{C}$, as in \Cref{prop:marginals}. 
We henceforth assume that each $x \in \mathcal{X}$ is naturally represented by exactly $d$ components, i.e., $\mathcal{X} \subseteq \binom{\mathcal{C}}{d}$. 
This is often the case for usual applications and benchmarks in the GFlowNet literature, such as set generation \cite{jang2024learning}, phylogenetic inference \citep{zhou2024phylogfn}, design of mRNA sequences \citep{laajil2025curriculumaugmentedgflownetsmrnasequence}, causal discovery \citep{experts}, and certain combinatorial optimization tasks \citep{zhang2023}.   
\looseness=-1 

\begin{figure}[!h]
    \centering
    \includegraphics[width=.65\linewidth, page=5]{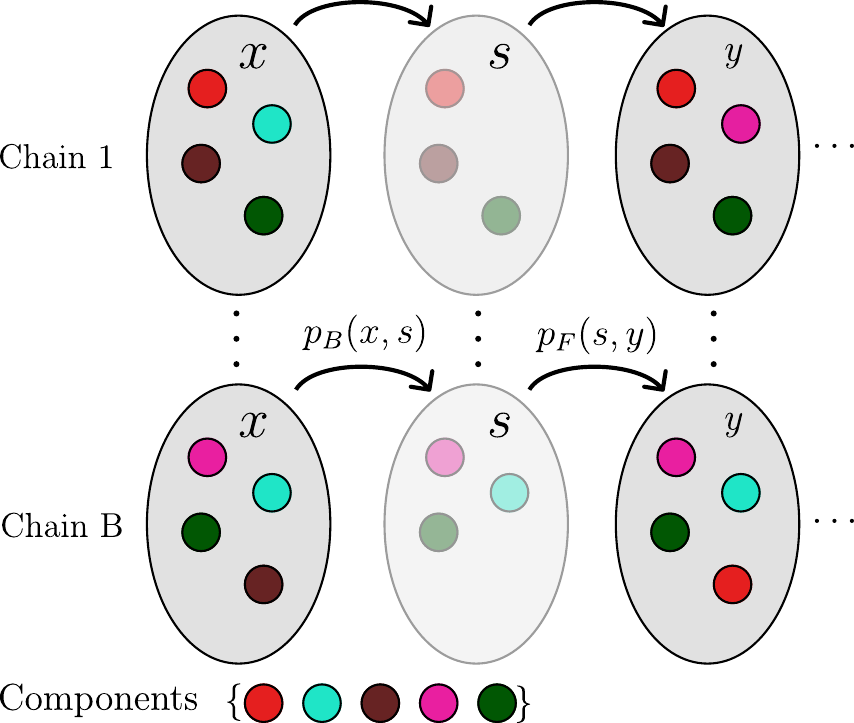}
    \caption{
        \textbf{A transition $x \rightarrow y$ for P-GFlowNets'} persistent Gibbs sampler. 
        At each iteration, we replace a component of the current state $x$ and evaluate the loss function in \Cref{eq:pgflownets} by averaging over the $B$ 
        stochastic processes. 
        This requires $\mathcal{O}(1)$ forward passess, regardless of $d$. 
        \looseness=-1 
    }
    \label{fig:pgflownets}
\end{figure}

To build intuition, consider \Cref{fig:pgflownets}. 
We start by generating samples $\{x_{t}^{b}\}_{b=1}^{B} \subseteq \mathcal{X}$ from an untrained GFlowNet. 
Each $x_{b}$ can be thought of as a $d$-sized subset of $\mathcal{C}$, for instance, $x_{t}^{b} = \{c_{1}^{t, b}, \dots, c_{d}^{t, b}\}$. 
At each iteration, we remove a component $c_{i}^{t, b}$ from $x_{t}^{b}$ according to $p_{B}(x_{t}^{b}, \cdot)$, resulting in $s_{t}^{b}$. 
We then choose a $c \in \mathcal{C}$ according to the forward policy $p_{F}(s_{t}^{b}, \cdot)$ and attach it to $s_{t}^{b}$, generating $x_{t + 1}^{b} \coloneqq s_{t}^{b} \cup \{c\}$.
In conclusion, we may update each model via a gradient step on \looseness=-1 
\begin{equation*}
    \hat{\mathcal{L}}_{\mathrm{C}} \coloneqq \frac{1}{B} \sum_{1 \le b \le B} \left( \log \frac{F(s_{t}^{b}) p_{F}(s_{t}^{b}, x_{t + 1}^{b})}{R(x_{t + 1}^{b}) p_{B}(x_{t + 1}^{b}, s_{t}^{b})} \right)^{2}. 
\end{equation*}
Upon repetition, this 
produces a coupled Markov chain $\{\{x_{t}^{b}\}_{b=1}^{B}\}_{t \ge 1}$ satisfying the DB condition for terminal ($x_{t}^{b}$) and near-terminal ($s_{t}^{b}$) states.
However, 
independent sampling via MDP simulation can only be achieved if the DB condition is satisfied for every intermediate state.
In view of this, we also sample $s_{t}^{b, k} \sim p_{B}^{k}(x_{t}^{b}, \cdot)$ 
by removing $k$ components from $x_{t}^{b}$; $p_{B}^{k}$ denotes $p_{B}$'s $k$-fold composition. 
Then, we select 
$s_{t + 1}^{b, k} \sim p_{F}(s_{t}^{b, k}, \cdot)$ and compute \looseness=-1 
\begin{equation*}
    \hat{\mathcal{L}}_{\mathrm{I}} \coloneqq \frac{1}{B} \sum_{1 \le b \le B} \left( \log \frac{F(s_{t}^{b, k}) p_{F}(s_{t}^{b, k}, s_{t + 1}^{b, k})}{F(s_{t + 1}^{b, k}) p_{B}(s_{t + 1}^{b, k}, s_{t}^{b, k})} \right)^{2}. 
\end{equation*}
In practice, we pick $k \sim \mathrm{Cat}(\mathbf{w})$ with $\mathbf{w} \in \mathbb{R}^{d}$ as the probabilities of a truncated Poisson distribution with average equal to $\log d$.
As suggested by \Cref{prop:dbcon}, this choice provides larger weight to near-terminal states, which has been shown to be beneficial \citep{silva2025when}, while ensuring coverage of the entire state graph. We then define
\begin{equation} \label{eq:pgflownets} 
    \hat{\mathcal{L}}_{\mathrm{P}} = \hat{\mathcal{L}}_{\mathrm{C}} + \hat{\mathcal{L}}_{\mathrm{I}} 
\end{equation}
as our loss function. 
We refer to a model trained by minimizing $\hat{\mathcal{L}}_{\mathrm{P}}$ as a \emph{Particle (P) GFlowNet}. 
Notably, we show below that the Markov chain $\{x_{t}\}_{t \ge 1}$ described above is ergodic with respect to the target $\pi$ when the P-GFlowNet abides by the DB condition. 
This ensures P-GFlowNets support both correlated and independent sampling, the choice of which to use being a trade off between computational cost with statistical efficiency. \looseness=-1 
\looseness=-1 


\begin{proposition} \label{prop:ergodic}
    Let $(p_{F}, p_{B}, F)$ be a P-GFlowNet abiding by the detailed balance condition. 
    Define $\{x_{t}\}_{t\ge1}$ as the Markov chain with the transition kernel 
    depicted in \Cref{fig:pgflownets}.
    Then, $\{x_{t}\}_{t\ge1}$
    is ergodic with respect to $\pi$. \looseness=-1 
\end{proposition}


Importantly, 
our persistent Gibbs chain might suffer from inadequate state space exploration 
due to the structural similarity of adjacent states. 
To mitigate this issue, we occasionally 
refresh the process with fresh independent samples from the current policy $p_{F}(s_{o}, \cdot)$. 
This approach 
is discussed next. \looseness=-1 






\paragraph{Chain rejuvenation.} 
We use the $\hat{R}$ metric to decide when to restart our stochastic process. 
Following standard statistical practice, we define $\hat{R} > 1.1$ as our condition for rejuvenation \citep{carpenter2017stan}.
To understand this, recall that the $\hat{R}$ measures the discrepancy between the within- and inter-chain variances.
When samples are independently generated, $\hat{R} \approx 1$. 
A large $\hat{R}$ indicates that the diversity of our batched stochastic process is significantly larger than that of each individual sequence, which suggests inefficient state space exploration.
To account for the non-stationarity of our Gibbs sampler, we use Gelman's split $\hat{R}$ metric. \looseness=-1  

As we are dealing with 
discrete state spaces, however, we cannot directly compute meaningful variances. 
Instead, we use the last layer embeddings of the forward policy's neural network to compute the split $\hat{R}$ metric.
As we show in the following section, the 
proposed criterion 
notoriously 
improves learning convergence and exploration. 
\looseness=-1






\section{Experiments} 
\label{sec:experiments} 

\begin{figure}
    \begin{subfigure}[t]{\linewidth} 
        \centering 
        \includegraphics[width=\linewidth]{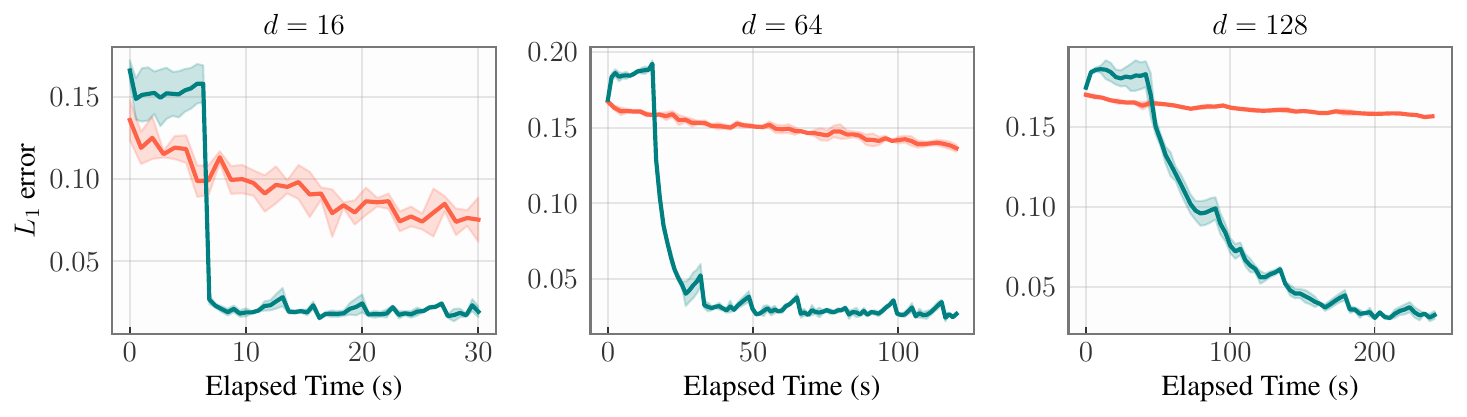}
        \caption{Set generation with log-additive rewards.}
    \end{subfigure}
    
    \begin{subfigure}[t]{\linewidth}
        \centering 
        \includegraphics[width=\linewidth]{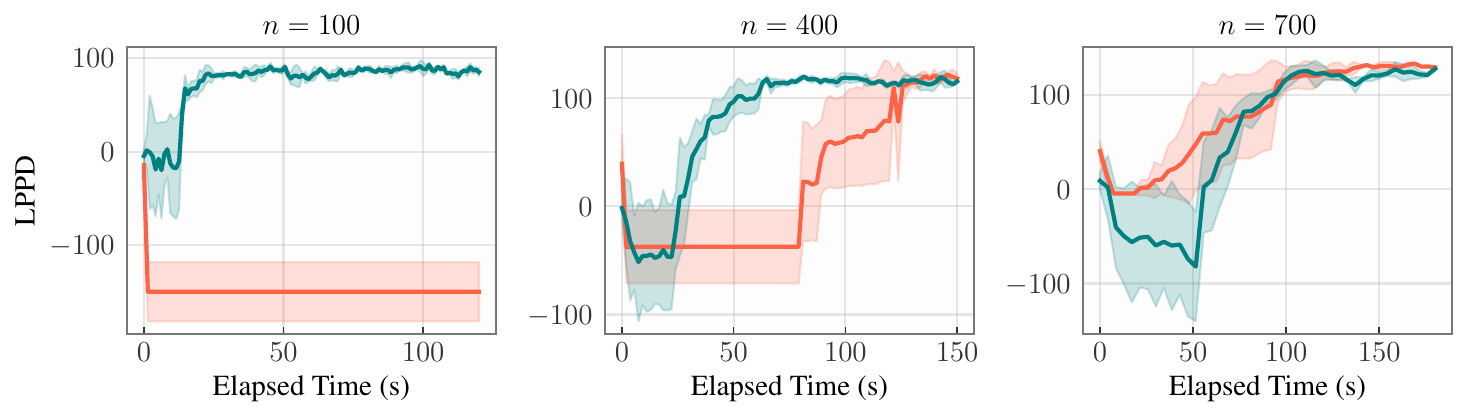}
        \caption{Bayesian variable selection.}
    \end{subfigure}
    
    \begin{subfigure}[t]{.6\linewidth}
        \centering
        \includegraphics[width=\linewidth]{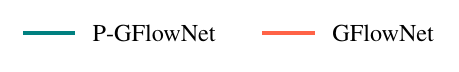}
    \end{subfigure}
    \caption{\textbf{P-GFlowNets often converge faster to the target distribution}.
    As the MDP horizon $d$ grows (a), the relative speed up of our method increases. 
    In contrast (b), our method grows more effective as the reward query cost decreases relative to the 
    sampling cost---i.e., as the number $n$ of samples for likelihood evaluation becomes smaller. \looseness=-1 
    }
    \label{fig:setsvarsel} 
\end{figure}

\begin{figure}
    \begin{subfigure}[t]{\linewidth}
        \centering
        \includegraphics[width=\linewidth]{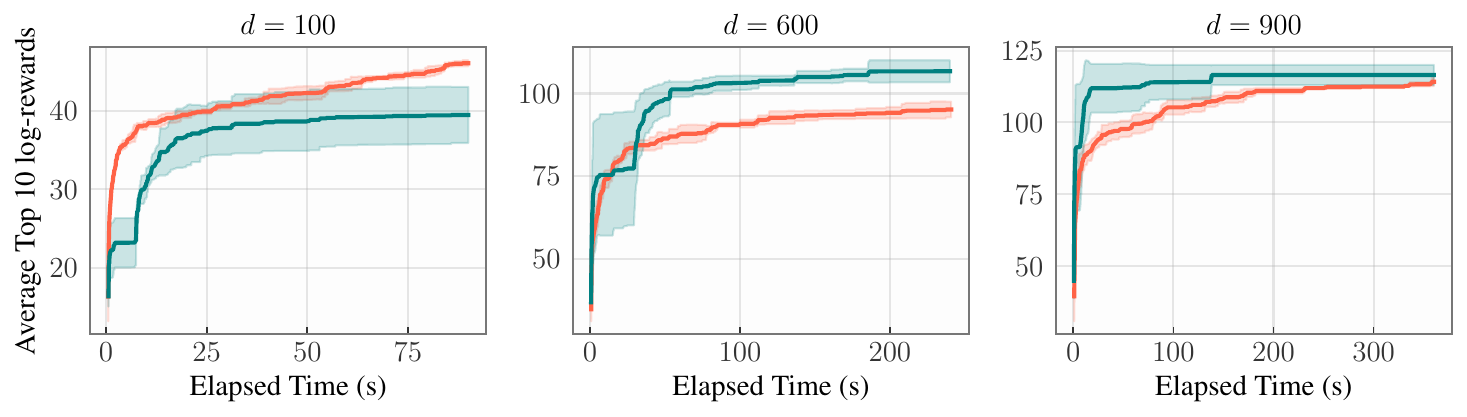} 
        \caption{Ising model.} 
    \end{subfigure}
    
    \begin{subfigure}[t]{\linewidth} 
        \centering
        \includegraphics[width=\linewidth]{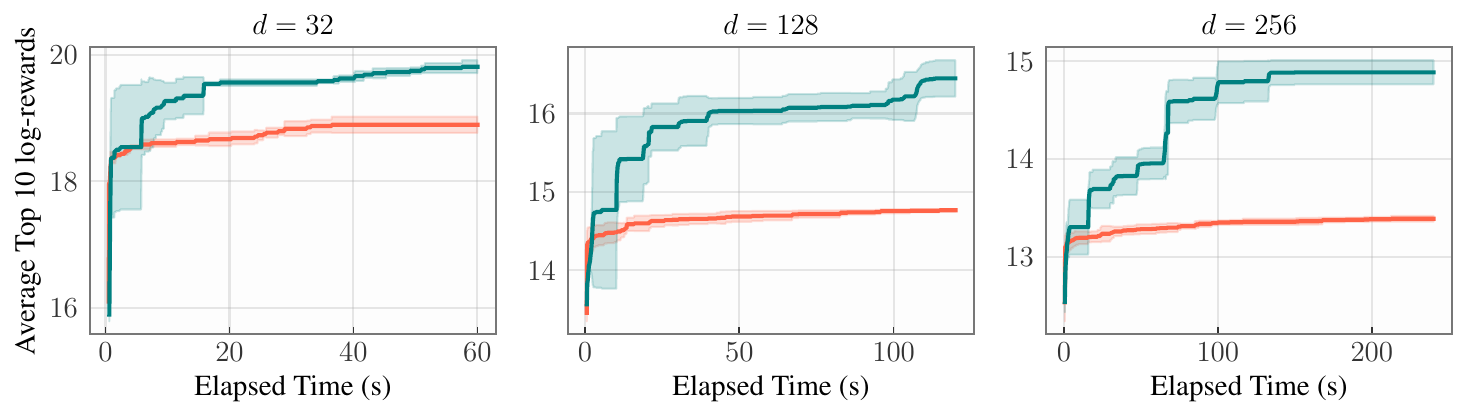}
        \caption{Bit sequences.} 
    \end{subfigure}
    
    \begin{subfigure}[t]{.6\linewidth}
        \centering
        \includegraphics[width=\linewidth]{uai2026-template/figures/legend_1x2.pdf}
    \end{subfigure}
    \caption{
        \textbf{P-GFlowNets improve exploration.} 
        We show
        the average log-reward of the $10$ most rewarding states found in 
        training.
        As in \Cref{fig:setsvarsel}, the 
        runtime gains from P-GFlowNets 
        as the MDP horizon ($d$) grows. \looseness=-1 
    }
    \label{fig:isingbits} 
    \vspace{-12pt} 
\end{figure}

Our experimental campaign addresses the following research questions (RQs) regarding P-GFlowNets.
\begin{enumerate}[leftmargin=23.9pt]
    \item[RQ1] Under which conditions do P-GFlowNets 
    improve 
    convergence relative to a standard GFlowNet? \looseness=-1  
    \item[RQ2] How effective is our approach for chain rejuvenation? 
\end{enumerate}
As P-GFlowNets reduce the number of neural network forward passes 
for sample generation, we expect it to reduce training time when learning is bottlenecked by policy evaluation (RQ1). 
This is the case when 
trajectories are long, and sampling is expensive, and 
reward queries are cheaper than multiple model inferences. 
Our experiments confirm this intuition (see \Cref{fig:isingbits,fig:setsvarsel}). 
\looseness=-1 

\begin{figure}
    \centering
    \includegraphics[width=\linewidth]{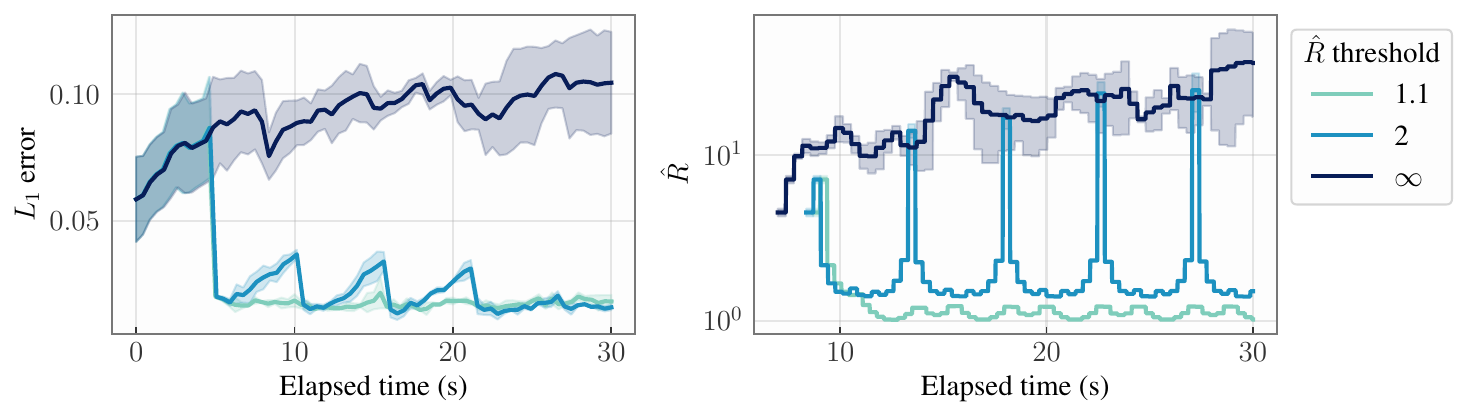}
    \caption{
    \textbf{Chain rejuvenation} critically accelerates learning. 
    Left: \Cref{eq:aaaaaa}. Right: $\hat{R}$ throughout training. 
    }
    \label{fig:AAA}
    \vspace{-12pt} 
\end{figure}

We also show that 
chain rejuvenation 
does not only improve state space exploration, but significantly accelerates learning convergence (RQ2). 
All in all, our observations position P-GFlowNets as a compute-efficient and principled algorithm for GFlowNet training. 
We provide further 
details and computer code for 
our experiments 
in the supplement.
\looseness=-1 

\paragraph{Set generation with log-additive rewards.} 
Our first task consists of generating fixed-size subsets of a given set $\mathcal{C} = \{1, \dots, K\}$.
Let $\mathcal{X} \coloneqq \{ s \subseteq \mathcal{C} \colon |s| = d\}$ be the space of $d$-sized subsets of $\mathcal{C}$.
Similarly, $\mathcal{S} \coloneqq \{s \subseteq \mathcal{C} \colon |s| < d\}$, and $s_{o} = \emptyset$ be the initial state.
Given a 
utility function $u \colon \mathcal{C} \rightarrow \mathbb{R}_{+}$, we define the target distribution 
$R \colon \mathcal{X} \rightarrow \mathbb{R}_{+}$ 
as \looseness=-1 
\begin{equation*}
    \log R(x) = \sum_{c \in x} \log u(c). 
\end{equation*}
This function has several important features. 
First, despite being factorizable into $x$'s components, it cannot be arbitrarily well-approximated by a mean-field variational approximation. 
Second, the function $R(x)$ can be naturally extended to arbitrary $d$ (state graph's diameter) and $K$ (state graph branching's factor) with $d \le K$.
Third, if $\mathbf{1}_{s}$ denotes $s$'s indicator function, both the partition function $Z \coloneqq \sum_{x \in \mathcal{X}} R(x)$ and the marginals $p_{c} \coloneqq \frac{1}{Z} \sum_{s \in \mathcal{X}} R(x) \cdot \mathbf{1}_{s}(c)$ for $c \in \mathcal{C}$ can be efficiently computed in $\mathcal{O}(K \cdot d^{2})$ through the following dynamic programming algorithm, which may be of independent interest for GFlowNet evaluation. 
Let $\mathrm{first}_{k}(S)$ denote the first $k$ elements of any $S \subseteq \mathcal{C}$ in a fixed order, and define \looseness=-1   
\begin{equation*}
    Z_{k, i}^{c} = \!\!\!\!\!\! \sum_{\substack{s \subseteq \mathrm{first}_{k}(\mathcal{C}\setminus \{c\}) \\ |s| = i}}  \prod_{e \in s} u(e) \text{ and } Z_{k, i} = \sum_{\substack{s \subseteq \mathrm{first}_{k}(\mathcal{C}) \\ |s| = i}} \prod_{e \in s} u(e) 
\end{equation*}
for $k \in \{1, \dots, K\}$ and $i \in \{1, \dots d\}$. 
It should be clear that if $P_{k, i}^{c} = u(c) \cdot \frac{Z_{k - 1, i}^{c}}{Z_{k, i}}$ then $p_{c} = P_{K, d}^{c}$. 
Also, $Z_{k + 1, i} = u(k + 1) \cdot Z_{k, i - 1} + Z_{k, i}$, with a similar recurrence equation for $Z_{k, i}^{c}$ obtained by replacing $u(k + 1)$ by the appropriate $(k + 1)$-th element of $\mathcal{C} \setminus \{c\}$. 
As such, we gauge the 
accuracy of a trained GFlowNet on this task by drawing 
$N$ independent states $\{\{x_{1}, \dots, x_{N}\}\} \subseteq \mathcal{X}$ 
and computing
\begin{equation} \label{eq:aaaaaa} 
    \frac{1}{|\mathcal{C}|} \sum_{c \in \mathcal{C}} |p_{c} - \hat{p}_{c}| \text{ with } \hat{p}_{c} = \frac{1}{N} \sum_{1 \le n \le N} \mathbf{1}_{x_{n}}(c).  
\end{equation}
We consider $(d, K) \in \{(32, 64), (64, 128), (128, 256)\}$ and $\log u(c) \sim \mathcal{N}(0, 1)$ 
drawn from a standard Gaussian for $c \in \mathcal{C}$.  
Then, we track the reduction in the above metric in terms of wall-clock time in \Cref{fig:setsvarsel}.
As $d$ increases, the runtime gap between P-GFlowNet and a standard GFlowNet widens. 

Additionally, \Cref{fig:AAA} shows 
our rejuvenation approach 
is paramount for 
speeding up training. 
There, we consider $(d, K) = (64, 128)$ and the conditions $\hat{R} > \alpha$ for $\alpha = 1.1$ (default), $\alpha = 2$, and $\alpha = \infty$ (i.e., no rejuvenation). \looseness=-1 


\paragraph{Bayesian variable selection.} 
When reward query costs offset the sampling overhead, we expect that allocating more compute per sample should speed up learning convergence. 
Indeed, this has been empirically observed by \citet{madan2025towards, kim2025adaptive, dallantonia2026avoidknowdivergenttrajectory}. 
To understand how this behavior affects the training of P-GFlowNets, we consider the problem of Bayesian variable selection \citep{george1993variable} with progressively larger sample sizes
corresponding to increasingly expensive-to-evaluate posterior distributions. 
Let $\mathbf{X} \in \mathbb{R}^{n \times K}$ and $\mathbf{y} \in \mathbb{R}^{n}$ be a dataset with $n$ samples and $K$-dimensional features, and denote by $\mathbf{X}_{F} \in \mathbb{R}^{n \times |F|}$ the column-filtered data with $F \subseteq \{1, \dots, K\}$. 
Also, let $\mathbf{I}_{n}$ (resp. $\mathbf{I}_{|F|}$) is $n$-dimensional (resp. $|F|$-dimensional) identity matrix, $F \sim \mathrm{Multinomial}(\{1, \dots, K\}, \pi)$ indicates each $F$ is sampled with probability $\pi^{|F|} (1 - \pi)^{K - |F|}$ for $\pi \in [0, 1]$. 
We then consider the linear model 
\looseness=-1 
\begin{equation*}
    \begin{split} 
        \mathbf{y} | \beta, F \sim \mathcal{N}(\mathbf{X}_{F} \beta_{F}, \eta^{2} \mathbf{I}_{n}) \text{ with } \\ \beta_{F} | F \sim \mathcal{N}(0, \nu^{2} \mathbf{I}_{|F|}) \text{ and } F \sim \mathrm{Multinomial}([K], \pi). 
    \end{split} 
\end{equation*}
We assume $\pi$, $\eta > 0$, and $\nu > 0$ are known.
Under these conditions, the marginal posterior distribution over the set $F$ is 
\begin{equation*}
    \log R(F) = \log \mathcal{N}(\mathbf{y} | 0, \nu^{2} \mathbf{X}_{F} \mathbf{X}_{F}^{T} + \eta^{2} \mathbf{I}_{n}) + |F| \log \frac{\pi}{1 - \pi}, 
\end{equation*}
with $\mathcal{N}(\mathbf{y} | \mu, \Sigma)$ denoting the density of a Gaussian distribution with mean $\mu$ and covariance 
$\Sigma$. 
Clearly, evaluating $R(F)$ costs $\mathcal{O}(K \cdot n^{2})$. 
To 
assess a GFlowNet, we evaluate the log-predictive posterior density (LPPD) of a held-out dataset $(\mathbf{X}^{\star}, \mathbf{y}^{\star})$ throughout training. 
Given samples $\{\{F_{1}, \dots, F_{N}\}\}$, 
we approximate the LPPD as 
\begin{equation*}
    \mathrm{LPPD} = \log \frac{1}{N} \sum_{1 \le n \le N} p(\mathbf{y}^{\star} | \mathbf{X}^{\star}, \mathbf{y}, F), 
\end{equation*}
with $p(\mathbf{y}^{\star} | \mathbf{y}, F) = \mathcal{N}(\mathbf{y}^{\star} | \mathbf{X}_{F} \mu_{F}, \mathbf{X}_{F}^{\star} \Sigma_{F} \mathbf{X}^{T} + \eta^{2} \mathbf{I}_{n})$ and $\mu_{F} = \eta^{-2} \Sigma_{F} \mathbf{X}_{F}^{T} \mathbf{y}$ and $\Sigma_{F} = (\eta^{-2} \mathbf{X}_{F}^{T} \mathbf{X}_{F} + \nu^{-2} \mathbf{I}_{|F|})^{-1}$ as $\beta_{F}$ posterior's mean and covariance given $F$, respectively. 
In practice, each row of $\mathbf{X}_{F}$ is independently sampled from $\mathcal{N}(\mathbf{0}, \Gamma)$, with $\mathbf{0} \in \mathbb{R}^{K}$ and $\Gamma \in \mathbb{R}^{K \times K}$ and $\Gamma_{ij} = \gamma^{|i - j|}$ for $\gamma = 0.8$.
Under this model, $\mathbf{X}$'s columns are highly correlated, and the posterior distribution over $F$ is multimodal. 

Notably, \Cref{fig:setsvarsel} shows that P-GFlowNets consistently outperform a standard GFlowNet when $n$ small.
However, as $n$ grows, the posterior evaluation cost outpaces that of trajectory sampling, and the computational benefits from our model dwindle. 
As explained above, this confirms our initial assumptions regarding P-GFlowNets. \looseness=-1 

\looseness=-1 

We next 
consider the tasks of Ising model simulation, also present in \citep{liu2023mam}, and of bit generation, 
a common testbed for GFlowNets \citep[e.g.,][]{torchgfn}. 
For each example, $\mathcal{X}$ is the space of $d$-sized sequences with elements from $\{-1, 1\}$ and $\{1, 0\}$, respectively. 
As is standard practice \cite{malkin2022trajectory, lingtrajectory, kim2025adaptive},
we evaluate a model by measuring the average log reward for the top $10$ most valuable samples found throughout 
training.  
\looseness=-1 



\paragraph{Ising model.} 
Simply put, let $\mathbf{J} \in \mathbb{R}^{d \times d}$ and $\mathbf{h} \in \mathbb{R}^{d}$. 
We define an energy function as $E(\mathbf{x}) = -\frac{1}{2} \mathbf{x}^{T} \mathbf{J} \mathbf{x} - \mathbf{h}^{T} \mathbf{x}$, and $p(x) \propto \exp\left\{-\nicefrac{E(x)}{\beta}\right\}$ as the probability of a configuration $x \in \{-1, 1\}^{d}$ under a temperature $\beta > 0$. 
In \Cref{fig:isingbits}, we consider $d \in \{100, 300, 900\}$. 
We observe that P-GFlowNets drastically improve exploration when trajectories are long and sampling is consequently expensive. 
\looseness=-1 

\paragraph{Bit sequences.} 
As in \citet{malkin2022trajectory, tiapkin2024generative}, we let $\mathcal{M} \subseteq \{1, 0\}^{d}$ be a set of \emph{modes}, and define $\beta \log R(x) = 1 - \min_{m \in \mathcal{M}} \nicefrac{\rho(m, x)}{d}$, with $\rho(x, m)$ 
as edit distance between $x$ and $m$ and $\beta > 0$ is a temperature parameter.
We consider $d \in \{64, 128, 256\}$ and $\beta = \nicefrac{1}{20}$. 
As we consistently observed in prior experiments, 
\Cref{fig:isingbits}
shows 
P-GFlowNets significantly speed up the discovery of high-reward states as the MDP horizon grows. 
\looseness=-1 

\section{Discussion} 
\label{sec:discussions} 

We showed that MaMs \citep{liu2023mam}, which were previously thought to be distinct from GFlowNets, can be seen as an instantiation of a conditional GFlowNet using a persistent Gibbs sampler for exploration during training. 
Based on this, we also demonstrated this strategy generalizes beyond 
autoregressive modelling, for which MaMs were originally designed, while maintaining its computational benefits.   
Our experiments highlighted that the resulting method, called Particle GFlowNets, significantly accelerated learning convergence in terms of wall-clock time when compared against conventional GFlowNet training algorithms. 
\looseness=-1 

From a broader perspective, our work (esp. \Cref{prop:dbcon,prop:ergodic}) strengthens the connection between GFlowNets and Markov chain methods, which was also formally studied by \citet{markovchains}. 
This raises several questions.
How to optimally decide when to rejuvenate the persistent Gibbs sampler? 
As noted in \cite{silva2025when}, diagnosing GFlowNets is 
strikingly difficult; 
can we draw inspirations from the MCMC literature to properly assess the distributional accuracy of GFlowNets?
Successful Markov samplers, such as 
Langevin dynamics-based methods \citep{Welling2011,girolami2011riemann}, rely on simulating a latent dynamics for each transition of the underlying stochastic process; is such a technique extensible to GFlowNets, and when does it accelerate training? 
We believe these to be interesting directions for future research. 
\looseness=-1


\begin{acknowledgements}
    DM acknowledges the support of the Fundação Carlos Chagas Filho de Amparo à Pesquisa do Estado do Rio de Janeiro (FAPERJ) (SEI-260003/020348/2025, SEI-260003/020694/2025) and the Conselho Nacional de Desenvolvimento Científico e Tecnológico (CNPq) (404336/2023-0, 305692/2025-9, 445170/2024-7).
\end{acknowledgements}

\bibliography{uai2026-template}

@InProceedings{	  bengio2021,
  author	= {Bengio, Emmanuel and Jain, Moksh and Korablyov, Maksym and
		  Precup, Doina and Bengio, Yoshua},
  booktitle	= {NeurIPS (NeurIPS)},
  title		= {Flow Network based Generative Models for Non-Iterative
		  Diverse Candidate Generation},
  year		= {2021}
}

@Article{	  carpenter2017stan,
  title		= {Stan: A probabilistic programming language},
  author	= {Carpenter, Bob and Gelman, Andrew and Hoffman, Matthew D
		  and Lee, Daniel and Goodrich, Ben and Betancourt, Michael
		  and Brubaker, Marcus and Guo, Jiqiang and Li, Peter and
		  Riddell, Allen},
  journal	= {Journal of statistical software},
  year		= {2017},
  publisher	= {Columbia Univ., New York, NY (United States); Harvard
		  Univ., Cambridge, MA (United States)}
}

@InProceedings{	  deleu2022bayesian,
  title		= {Bayesian Structure Learning with Generative Flow
		  Networks},
  author	= {Deleu, Tristan and G{\'o}is, Ant{\'o}nio and Emezue, Chris
		  Chinenye and Rankawat, Mansi and Lacoste-Julien, Simon and
		  Bauer, Stefan and Bengio, Yoshua},
  booktitle	= {UAI},
  year		= {2022}
}

@InProceedings{	  deleu2023joint,
  title		= {Joint {Bayesian} Inference of Graphical Structure and
		  Parameters with a Single Generative Flow Network},
  author	= {Deleu, Tristan and Nishikawa-Toomey, Mizu and Subramanian,
		  Jithendaraa and Malkin, Nikolay and Charlin, Laurent and
		  Bengio, Yoshua},
  booktitle	= {Advances in Neural Processing Systems (NeurIPS)},
  year		= {2023}
}

@Article{	  foundations,
  author	= {Yoshua Bengio and Salem Lahlou and Tristan Deleu and
		  Edward J. Hu and Mo Tiwari and Emmanuel Bengio},
  title		= {GFlowNet Foundations},
  journal	= {Journal of Machine Learning Research (JMLR)},
  year		= {2023}
}

@Article{	  geman1984,
  doi		= {10.1109/tpami.1984.4767596},
  year		= {1984},
  month		= nov,
  publisher	= {Institute of Electrical and Electronics Engineers
		  ({IEEE})},
  volume	= {{PAMI}-6},
  number	= {6},
  author	= {Stuart Geman and Donald Geman},
  title		= {Stochastic Relaxation, Gibbs Distributions, and the
		  Bayesian Restoration of Images},
  journal	= {{IEEE} Transactions on Pattern Analysis and Machine
		  Intelligence}
}

@Misc{		  hu2023amortizing,
  title		= {Amortizing intractable inference in large language
		  models},
  author	= {Edward J. Hu and Moksh Jain and Eric Elmoznino and
		  Younesse Kaddar and et al.},
  year		= {2023},
  eprint	= {2310.04363},
  archiveprefix	= {arXiv},
  primaryclass	= {cs.LG}
}

@Article{	  jain2023gflownets,
  title		= {Gflownets for ai-driven scientific discovery},
  author	= {Jain, Moksh and Deleu, Tristan and Hartford, Jason and
		  Liu, Cheng-Hao and Hernandez-Garcia, Alex and Bengio,
		  Yoshua},
  journal	= {Digital Discovery},
  year		= {2023},
  publisher	= {Royal Society of Chemistry}
}

@InProceedings{	  jang2024learning,
  title		= {Learning Energy Decompositions for Partial Inference in
		  {GF}lowNets},
  author	= {Hyosoon Jang and Minsu Kim and Sungsoo Ahn},
  booktitle	= {The Twelfth International Conference on Learning
		  Representations},
  year		= {2024}
}

@Article{	  kim2023local,
  title		= {Local search gflownets},
  author	= {Kim, Minsu and Yun, Taeyoung and Bengio, Emmanuel and
		  Zhang, Dinghuai and Bengio, Yoshua and Ahn, Sungsoo and
		  Park, Jinkyoo},
  journal	= {arXiv preprint arXiv:2310.02710},
  year		= {2023}
}

@Misc{		  kim2024localsearchgflownets,
  title		= {Local Search GFlowNets},
  author	= {Minsu Kim and Taeyoung Yun and Emmanuel Bengio and
		  Dinghuai Zhang and Yoshua Bengio and Sungsoo Ahn and
		  Jinkyoo Park},
  year		= {2024},
  eprint	= {2310.02710},
  archiveprefix	= {arXiv},
  primaryclass	= {cs.LG},
  url		= {https://arxiv.org/abs/2310.02710}
}

@Misc{		  lau2023dgfndoublegenerativeflow,
  title		= {DGFN: Double Generative Flow Networks},
  author	= {Elaine Lau and Nikhil Vemgal and Doina Precup and Emmanuel
		  Bengio},
  year		= {2023},
  eprint	= {2310.19685},
  archiveprefix	= {arXiv},
  primaryclass	= {cs.LG},
  url		= {https://arxiv.org/abs/2310.19685}
}

@InProceedings{	  lingtrajectory,
  title		= {Better Training of {GF}low{N}ets with Local Credit and
		  Incomplete Trajectories},
  author	= {Pan, Ling and Malkin, Nikolay and Zhang, Dinghuai and
		  Bengio, Yoshua},
  booktitle	= {International Conference on Machine Learning (ICML)},
  year		= {2023}
}

@InProceedings{	  madan2022learninggf,
  title		= {Learning GFlowNets from partial episodes for improved
		  convergence and stability},
  author	= {Kanika Madan and Jarrid Rector-Brooks and Maksym Korablyov
		  and Emmanuel Bengio and Moksh Jain and Andrei Cristian Nica
		  and Tom Bosc and Yoshua Bengio and Nikolay Malkin},
  booktitle	= {International Conference on Machine Learning},
  year		= {2022}
}

@InProceedings{	  malkin2022trajectory,
  title		= {Trajectory balance: Improved credit assignment in
		  {GF}lowNets},
  author	= {Nikolay Malkin and Moksh Jain and Emmanuel Bengio and Chen
		  Sun and Yoshua Bengio},
  booktitle	= {NeurIPS (NeurIPS)},
  year		= {2022}
}

@Article{	  malkin2023gflownets,
  title		= {{GFlowNets} and variational inference},
  author	= {Malkin, Nikolay and Lahlou, Salem and Deleu, Tristan and
		  Ji, Xu and Hu, Edward and Everett, Katie and Zhang,
		  Dinghuai and Bengio, Yoshua},
  journal	= {International Conference on Learning Representations
		  (ICLR)},
  year		= {2023}
}

@Misc{		  markovchains,
  title		= {Generative Flow Networks: a Markov Chain Perspective},
  author	= {Tristan Deleu and Yoshua Bengio},
  year		= {2023},
  eprint	= {2307.01422},
  archiveprefix	= {arXiv},
  primaryclass	= {cs.LG}
}

@Book{		  mcbook,
  author	= {Art B. Owen},
  year		= 2013,
  title		= {Monte Carlo theory, methods and examples}
}

@InProceedings{	  mogfn,
  title		= {Multi-Objective {GF}low{N}ets},
  author	= {Jain, Moksh and Raparthy, Sharath Chandra and
		  Hernandez-Garcia, Alex and Rector-Brooks, Jarrid and
		  Bengio, Yoshua and Miret, Santiago and Bengio, Emmanuel},
  booktitle	= {International Conference on Machine Learning (ICML)},
  year		= {2023}
}

@InProceedings{	  pan2024pretraining,
  title		= {Pre-Training and Fine-Tuning Generative Flow Networks},
  author	= {Ling Pan and Moksh Jain and Kanika Madan and Yoshua
		  Bengio},
  booktitle	= {The Twelfth International Conference on Learning
		  Representations},
  year		= {2024}
}

@InProceedings{	  robust,
  title		= {Robust scheduling with GFlowNets},
  author	= {Zhang, David W and Rainone, Corrado and Peschl, Markus and
		  Bondesan, Roberto},
  booktitle	= {International Conference on Learning Representations
		  (ICLR)},
  year		= {2023}
}

@Article{	  roy2023goal,
  title		= {Goal-conditioned gflownets for controllable
		  multi-objective molecular design},
  author	= {Roy, Julien and Bacon, Pierre-Luc and Pal, Christopher and
		  Bengio, Emmanuel},
  journal	= {arXiv preprint arXiv:2306.04620},
  year		= {2023}
}

@Book{		  schervish2012theory,
  title		= {Theory of statistics},
  author	= {Schervish, Mark J},
  year		= {2012},
  publisher	= {Springer Science \& Business Media}
}

@InProceedings{	  sequence,
  title		= {Biological Sequence Design with {GF}low{N}ets},
  author	= {Jain, Moksh and Bengio, Emmanuel and Hernandez-Garcia,
		  Alex and Rector-Brooks, Jarrid and Dossou, Bonaventure F.
		  P. and Ekbote, Chanakya Ajit and Fu, Jie and Zhang, Tianyu
		  and Kilgour, Michael and Zhang, Dinghuai and Simine, Lena
		  and Das, Payel and Bengio, Yoshua},
  booktitle	= {International Conference on Machine Learning (ICML)},
  year		= {2022}
}

@InProceedings{	  shen23gflownets,
  author	= {Shen, Max W. and Bengio, Emmanuel and Hajiramezanali,
		  Ehsan and Loukas, Andreas and Cho, Kyunghyun and
		  Biancalani, Tommaso},
  title		= {Towards Understanding and Improving GFlowNet Training},
  year		= {2023},
  booktitle	= {International Conference on Machine Learning}
}

@InProceedings{	  theory,
  author	= {Salem Lahlou and Tristan Deleu and Pablo Lemos and
		  Dinghuai Zhang and Alexandra Volokhova and Alex
		  Hern{\'{a}}ndez{-}Garc{\'{\i}}a and L{\'{e}}na N{\'{e}}hale
		  Ezzine and Yoshua Bengio and Nikolay Malkin},
  title		= {A theory of continuous generative flow networks},
  booktitle	= {{ICML}},
  series	= {Proceedings of Machine Learning Research},
  volume	= {202},
  pages		= {18269--18300},
  publisher	= {{PMLR}},
  year		= {2023}
}

@Misc{		  tiapkin2024generative,
  title		= {Generative Flow Networks as Entropy-Regularized RL},
  author	= {Daniil Tiapkin and Nikita Morozov and Alexey Naumov and
		  Dmitry Vetrov},
  year		= {2024},
  eprint	= {2310.12934},
  archiveprefix	= {arXiv},
  primaryclass	= {cs.LG}
}

@Misc{		  venkatraman2024amortizingintractableinferencediffusion,
  title		= {Amortizing intractable inference in diffusion models for
		  vision, language, and control},
  author	= {Siddarth Venkatraman and Moksh Jain and Luca Scimeca and
		  Minsu Kim and Marcin Sendera and Mohsin Hasan and Luke Rowe
		  and Sarthak Mittal and Pablo Lemos and Emmanuel Bengio and
		  Alexandre Adam and Jarrid Rector-Brooks and Yoshua Bengio
		  and Glen Berseth and Nikolay Malkin},
  year		= {2024},
  eprint	= {2405.20971},
  archiveprefix	= {arXiv},
  primaryclass	= {cs.LG},
  url		= {https://arxiv.org/abs/2405.20971}
}

@InProceedings{	  zhang2023,
  author	= {Dinghuai Zhang and Hanjun Dai and Nikolay Malkin and Aaron
		  Courville and Yoshua Bengio and Ling Pan},
  booktitle	= {NeurIPS (NeurIPS)},
  title		= {Let the Flows Tell: Solving Graph Combinatorial
		  Optimization Problems with GFlowNets},
  year		= {2023}
}

@Misc{		  zhou2023phylogfn,
  title		= {PhyloGFN: Phylogenetic inference with generative flow
		  networks},
  author	= {Mingyang Zhou and Zichao Yan and Elliot Layne and Nikolay
		  Malkin and Dinghuai Zhang and Moksh Jain and Mathieu
		  Blanchette and Yoshua Bengio},
  year		= {2023},
  eprint	= {2310.08774},
  archiveprefix	= {arXiv},
  primaryclass	= {q-bio.PE}
}

@InProceedings{	  zhou2024phylogfn,
  title		= {Phylo{GFN}: Phylogenetic inference with generative flow
		  networks},
  author	= {Ming Yang Zhou and Zichao Yan and Elliot Layne and Nikolay
		  Malkin and Dinghuai Zhang and Moksh Jain and Mathieu
		  Blanchette and Yoshua Bengio},
  booktitle	= {The Twelfth International Conference on Learning
		  Representations},
  year		= {2024}
}

@Misc{		  zhu2023sampleefficient,
  title		= {Sample-efficient Multi-objective Molecular Optimization
		  with GFlowNets},
  author	= {Yiheng Zhu and Jialu Wu and Chaowen Hu and Jiahuan Yan and
		  Chang-Yu Hsieh and Tingjun Hou and Jian Wu},
  year		= {2023},
  eprint	= {2302.04040},
  archiveprefix	= {arXiv},
  primaryclass	= {cs.LG}
}

@software{jax2018github,
  author = {James Bradbury and Roy Frostig and Peter Hawkins and Matthew James Johnson and Chris Leary and Dougal Maclaurin and George Necula and Adam Paszke and Jake Vander{P}las and Skye Wanderman-{M}ilne and Qiao Zhang},
  title = {{JAX}: composable transformations of {P}ython+{N}um{P}y programs},
  version = {0.3.13},
  year = {2018},
}

@InProceedings{liu2023mam,
  title={Generative Marginalization Models},
  author={Liu, Sulin and Ramadge, Peter J and Adams, Ryan P},
  booktitle={{ International Conference on Machine Learning (ICML) }},
  year={2024}
}

@inproceedings{Welling2011,
  author    = {Welling, Max and Teh, Yee Whye},
  title     = {Bayesian Learning via Stochastic Gradient Langevin Dynamics},
  booktitle = {Proceedings of the 28th International Conference on Machine Learning (ICML-11)},
  year      = {2011},
}

@article{GelmanRubin1992,
  author  = {Gelman, Andrew and Rubin, Donald B.},
  title   = {Inference from Iterative Simulation Using Multiple Sequences},
  journal = {Statistical Science},
  year    = {1992},
}

@article{girolami2011riemann,
  title={Riemann manifold Langevin and Hamiltonian Monte Carlo methods},
  author={Girolami, Mark and Calderhead, Ben},
  journal={Journal of the Royal Statistical Society: Series B (Statistical Methodology)},
  year={2011},
  publisher={Wiley Online Library},
}

@book{meyn2009markov,
  title={Markov Chains and Stochastic Stability},
  author={Meyn, Sean P and Tweedie, Richard L},
  year={2009},
  publisher={Cambridge University Press},
  address={Cambridge},
  edition={2nd},
  note={Cambridge Mathematical Library},
}

@inproceedings{
    madan2025towards,
    title={Towards Improving Exploration through Sibling Augmented {GF}lowNets},
    author={Kanika Madan and Alex Lamb and Emmanuel Bengio and Glen Berseth and Yoshua Bengio},
    booktitle={The Thirteenth International Conference on Learning Representations},
    year={2025},
}

@inproceedings{
    kim2025adaptive,
    title={Adaptive teachers for amortized samplers},
    author={Minsu Kim and Sanghyeok Choi and Taeyoung Yun and Emmanuel Bengio and Leo Feng and Jarrid Rector-Brooks and Sungsoo Ahn and Jinkyoo Park and Nikolay Malkin and Yoshua Bengio},
    booktitle={The Thirteenth International Conference on Learning Representations},
    year={2025},
    url={https://openreview.net/forum?id=BdmVgLMvaf}
}

@misc{malek2025lossguidedauxiliaryagentsovercoming,
      title={Loss-Guided Auxiliary Agents for Overcoming Mode Collapse in GFlowNets}, 
      author={Idriss Malek and Aya Laajil and Abhijith Sharma and Eric Moulines and Salem Lahlou},
      year={2025},
      eprint={2505.15251},
      archivePrefix={arXiv},
      primaryClass={cs.LG},
      url={https://arxiv.org/abs/2505.15251}, 
}

@misc{laajil2025curriculumaugmentedgflownetsmrnasequence,
      title={Curriculum-Augmented GFlowNets For mRNA Sequence Generation}, 
      author={Aya Laajil and Abduragim Shtanchaev and Sajan Muhammad and Eric Moulines and Salem Lahlou},
      year={2025},
      eprint={2510.03811},
      archivePrefix={arXiv},
      primaryClass={cs.LG},
      url={https://arxiv.org/abs/2510.03811}, 
}

@misc{dallantonia2026avoidknowdivergenttrajectory,
      title={Avoid What You Know: Divergent Trajectory Balance for GFlowNets}, 
      author={Pedro Dall'Antonia and Tiago da Silva and Daniel Csillag and Salem Lahlou and Diego Mesquita},
      year={2026},
      eprint={2602.17827},
      archivePrefix={arXiv},
      primaryClass={cs.LG},
      url={https://arxiv.org/abs/2602.17827}, 
}

@inproceedings{
    zhang2025hybridbalance,
    title={Hybrid-Balance {GF}lowNet for Solving Vehicle Routing Problems},
    author={Ni Zhang and Zhiguang Cao},
    booktitle={The Thirty-ninth Annual Conference on Neural Information Processing Systems},
    year={2025},
}

@misc{lew2022pcleanbayesiandatacleaning,
      title={PClean: Bayesian Data Cleaning at Scale with Domain-Specific Probabilistic Programming}, 
      author={Alexander K. Lew and Monica Agrawal and David Sontag and Vikash K. Mansinghka},
      year={2022},
      eprint={2007.11838},
      archivePrefix={arXiv},
      primaryClass={cs.LG},
      url={https://arxiv.org/abs/2007.11838}, 
}

@misc{lew2022recursivemontecarlovariational,
      title={Recursive Monte Carlo and Variational Inference with Auxiliary Variables}, 
      author={Alexander K. Lew and Marco Cusumano-Towner and Vikash K. Mansinghka},
      year={2022},
      eprint={2203.02836},
      archivePrefix={arXiv},
      primaryClass={cs.LG},
      url={https://arxiv.org/abs/2203.02836}, 
}

@article{torchgfn,
  title={torchgfn: A PyTorch GFlowNet library},
  author={Viviano, Joseph D and Younis, Omar G and Choi, Sanghyeok and Schmidt, Victor and Bengio, Yoshua and Lahlou, Salem},
  journal={arXiv e-prints},
  pages={arXiv--2305},
  year={2023}
}

@misc{gritsaev2025optimizingbackwardpoliciesgflownets,
      title={Optimizing Backward Policies in GFlowNets via Trajectory Likelihood Maximization}, 
      author={Timofei Gritsaev and Nikita Morozov and Sergey Samsonov and Daniil Tiapkin},
      year={2025},
      eprint={2410.15474},
      archivePrefix={arXiv},
      primaryClass={cs.LG},
      url={https://arxiv.org/abs/2410.15474}, 
}

@inproceedings{
silva2025generalization,
title={Generalization and Distributed Learning of {GF}lowNets},
author={Tiago Silva and Amauri H Souza and Omar Rivasplata and Vikas Garg and Samuel Kaski and Diego Mesquita},
booktitle={The Thirteenth International Conference on Learning Representations},
year={2025},
}

@inproceedings{
silva2025when,
title={When do {GF}lowNets learn the right distribution?},
author={Tiago Silva and Rodrigo Barreto Alves and Eliezer de Souza da Silva and Amauri H Souza and Vikas Garg and Samuel Kaski and Diego Mesquita},
booktitle={The Thirteenth International Conference on Learning Representations},
year={2025},
}

@book{gibbs1902elementary,
  title={Elementary Principles in Statistical Mechanics: Developed with Especial Reference to the Rational Foundation of Thermodynamics},
  author={Gibbs, Josiah Willard},
  year={1902},
  publisher={Charles Scribner's Sons},
  address={New York},
  note={Reprinted by Dover Publications (1960) and others}
}

@article{hinton2002,
    author = {Hinton, Geoffrey E.},
    title = {Training products of experts by minimizing contrastive divergence},
    year = {2002},
    issue_date = {August 2002},
    publisher = {MIT Press},
    address = {Cambridge, MA, USA},
    volume = {14},
    number = {8},
    issn = {0899-7667},
    url = {https://doi.org/10.1162/089976602760128018},
    doi = {10.1162/089976602760128018},
    journal = {Neural Comput.},
    month = aug,
    pages = {1771–1800},
    numpages = {30}
}

@inproceedings{tieleman2008training,
  title={Training restricted Boltzmann machines using approximations to the likelihood gradient},
  author={Tieleman, Tijmen},
  booktitle={Proceedings of the 25th international conference on Machine learning},
  pages={1064--1071},
  year={2008}
}

@article{experts,
    author = {Silva, Tiago and Bazaluk, Bruna and da Silva, Eliezer and Góis, António and Lahlou, Salem and Heider, Dominik and Kaski, Samuel and Mesquita, Diego and Ribeiro, Adele},
    year = {2026},
    month = {01},
    journal = {SSRN},
    pages = {},
    title = {Expert-Aided Causal Discovery of Ancestral Graphs},
    doi = {10.2139/ssrn.6074306}
}

@article{george1993variable,
  title={Variable selection via Gibbs sampling},
  author={George, Edward I and McCulloch, Robert E},
  journal={Journal of the American Statistical Association},
  volume={88},
  number={423},
  pages={881--889},
  year={1993},
  publisher={Taylor \& Francis}
}

@misc{vaswani2023attentionneed,
      title={Attention Is All You Need}, 
      author={Ashish Vaswani and Noam Shazeer and Niki Parmar and Jakob Uszkoreit and Llion Jones and Aidan N. Gomez and Lukasz Kaiser and Illia Polosukhin},
      year={2023},
      eprint={1706.03762},
      archivePrefix={arXiv},
      primaryClass={cs.CL},
      url={https://arxiv.org/abs/1706.03762}, 
}

@misc{choi2025reinforcedsequentialmontecarlo,
      title={Reinforced sequential Monte Carlo for amortised sampling}, 
      author={Sanghyeok Choi and Sarthak Mittal and Víctor Elvira and Jinkyoo Park and Nikolay Malkin},
      year={2025},
      eprint={2510.11711},
      archivePrefix={arXiv},
      primaryClass={cs.LG},
      url={https://arxiv.org/abs/2510.11711}, 
}

@misc{hu2024squarederrorexploringloss,
      title={Beyond Squared Error: Exploring Loss Design for Enhanced Training of Generative Flow Networks}, 
      author={Rui Hu and Yifan Zhang and Zhuoran Li and Longbo Huang},
      year={2024},
      eprint={2410.02596},
      archivePrefix={arXiv},
      primaryClass={cs.LG},
      url={https://arxiv.org/abs/2410.02596}, 
}

@misc{jordan2024muon,
  author       = {Keller Jordan and Yuchen Jin and Vlado Boza and You Jiacheng and
                  Franz Cesista and Laker Newhouse and Jeremy Bernstein},
  title        = {Muon: An optimizer for hidden layers in neural networks},
  year         = {2024},
  url          = {https://kellerjordan.github.io/posts/muon/}
}

@misc{hu2023gflownetemlearningcompositionallatent,
      title={GFlowNet-EM for learning compositional latent variable models}, 
      author={Edward J. Hu and Nikolay Malkin and Moksh Jain and Katie Everett and Alexandros Graikos and Yoshua Bengio},
      year={2023},
      eprint={2302.06576},
      archivePrefix={arXiv},
      primaryClass={cs.LG},
      url={https://arxiv.org/abs/2302.06576}, 
}

@article{fitch1971toward,
  title={Toward defining the course of evolution: minimum change for a specific tree topology},
  author={Fitch, Walter M.},
  journal={Systematic Zoology},
  volume={20},
  number={4},
  pages={406--416},
  year={1971},
  publisher={Oxford University Press}
}

\newpage

\appendix 

\onecolumn

\title{Particle GFlowNets: Rethinking Generative Marginalization Models\\(Supplementary Material)}
\maketitle


\section{Summary \& Pseudocode}
\label{sec:summaryas} 

\begin{table}[!ht]
  \centering
  \caption{Summary of our main theoretical results.}
  \label{tab:resultsa}
  \begin{tabular}{@{}l l p{0.55\textwidth}@{}}
    \toprule
    \textbf{Result} & \textbf{Name} & \textbf{Statement} \\
    \midrule
    Prop.~\ref{prop:marginals} & Flow $=$ Marginal & The flow at any partial state equals the sum of rewards over all completions consistent with that state. \\[4pt]
    Cor.~\ref{col:marginals} & $\sigma$-Independence of Flow & The optimal flow does not depend on the variable ordering $\sigma$ used to construct the DAG. \\[4pt]
    Cor.~\ref{col:conditionals} & $\sigma$-Independence of Policy & The forward policy is likewise permutation-independent, so the network need not receive $\sigma$ as input. \\[4pt]
    Prop.~\ref{prop:equivalences} & MaMs $\equiv$ PC-GFlowNets & Consistent MaMs and PC-GFlowNets satisfying detailed balance are equivalent. \\[4pt]
    Prop.~\ref{prop:dbcon} & Consistency Error $=$ DB Loss & The MaM consistency error is an unbiased, $O(1)$-cost estimator of the weighted detailed balance training loss. \\[4pt]
    Prop.~\ref{prop:ergodic} & P-GFlowNet Ergodicity & The persistent Markov chain induced by a P-GFlowNet satisfying detailed balance converges to $\pi \propto R$. \\
    \bottomrule
  \end{tabular}
\end{table}

The main purpose of our work is to show that MaMs and GFlowNets, previously thought to be distinct, are the same---with MaM differing from conventional GFlowNet implementations solely in how states are sampled during training. 
Building on this equivalence, we introduce P-GFlowNets, demonstrating its correctness and computational efficiency as a sampling model for discrete, compositional spaces. 
In this context, \Cref{tab:resultsa} summarizes our central claims regarding the equivalence between P-GFlowNets and MaMs (\Cref{prop:marginals,prop:equivalences}, \Cref{col:marginals,col:conditionals}) and the correctness of P-GFlowNets (\Cref{prop:dbcon,prop:ergodic}).
We also provide a pseudocode description of P-GFlowNets in Algorithms~\ref{alg:transitionsa} and~\ref{alg:pgfn}. 

\begin{algorithm}[H]
\caption{Transition kernel for Algorithm~\ref{alg:pgfn}.}
\label{alg:transitionsa} 
\setcounter{AlgoLine}{0}
\Proc{\normalfont TransitionKernel$(x)$}{
  \tcc{One step of the persistent Gibbs chain (used to produce $y^{(b)}$ in $\Lc$)}
  Sample $s \sim p_B(\cdot \mid x)$\; \tcp*[f]{remove one component via backward policy}\\
  Sample $x' \sim p_F(\cdot \mid s)$\; \tcp*[f]{add one component via forward policy}\\
  \Return $x'$\;
}
\end{algorithm}

\begin{algorithm}[H]
\caption{P-GFlowNet Training Loop}
\label{alg:pgfn}
\KwIn{Reward function $R$, batch size $B$, GR threshold $\Rthr$, weight vector $\mathbf{w}$ (truncated Poisson, mean $\log d$)}
\KwOut{Trained parameters $\theta = (p_F, p_B, F)$}
\tcc{Initialisation}
Sample particles $\{x^{(b)}\}_{b=1}^{B} \sim p_F(\cdot \mid s_o)$ via MDP simulation\;
\While{\normalfont not converged}{
  \tcc{Complete loss $\Lc$ (terminal / near-terminal transitions)}
  \For{$b = 1$ \KwTo $B$}{
    $s^{(b)} \sim p_B(\cdot \mid x^{(b)})$\; \tcp*[f]{remove one component}\\
    $y^{(b)} \sim p_F(\cdot \mid s^{(b)})$\; \tcp*[f]{add one component}
  }
  $\Lc \gets \dfrac{1}{B} \sum_{b=1}^{B} \left( \log \dfrac{F(s^{(b)})\, p_F(s^{(b)}, y^{(b)})}{R(y^{(b)})\, p_B(y^{(b)}, s^{(b)})} \right)^{2}$\;
  \tcc{Intermediate loss $\Li$ (interior transitions, depth sampled from $\mathbf{w}$)}
  \For{$b = 1$ \KwTo $B$}{
    Sample depth $k^{(b)} \sim \mathrm{Cat}(\mathbf{w})$\; \tcp*[f]{truncated Poisson, mean $\log d$}\\
    $s^{(b,k)} \sim p_B^{k^{(b)}}(\cdot \mid x^{(b)})$\; \tcp*[f]{remove $k$ components}\\
    $s'^{(b,k)} \sim p_F(\cdot \mid s^{(b,k)})$\; \tcp*[f]{add one component}
  }
  $\Li \gets \dfrac{1}{B} \sum_{b=1}^{B} \left( \log \dfrac{F(s^{(b,k)})\, p_F(s^{(b,k)}, s'^{(b,k)})}{F(s'^{(b,k)})\, p_B(s'^{(b,k)}, s^{(b,k)})} \right)^{2}$\;
  \tcc{Gradient step}
  $\Lp \gets \Lc + \Li$\;
  $\theta \gets \theta - \eta \nabla_\theta \Lp(\theta)$\;
  \tcc{Particle update via Gibbs step}
  \For{$b = 1$ \KwTo $B$}{
    $x^{(b)} \gets y^{(b)}$\; \tcp*[f]{advance persistent chain}
  }
  \tcc{Rejuvenation (Gelman--Rubin criterion)}
  Compute split $\Rhat$ from last-layer embeddings of $p_F$ over the 512 most recent states\;
  \If{$\Rhat > \Rthr$}{
    Resample all particles: $x^{(b)} \sim p_F(\cdot \mid s_o)$ for $b = 1, \dots, B$\;
  }
}
\Return $\theta$\;
\end{algorithm}

\section{Proofs} 

\subsection{Proof of Proposition~\ref{prop:marginals}}


We first show that $F^{\sigma}(s_{o})$ equals the partition function of $Z$ and, in particular, does not depend on $\sigma$.
To see this, notice that \Cref{eq:dbsigma} implies 
\begin{equation} \label{eq:tbindb} 
    F^{\sigma} (s_{o}) p_{F}^{\sigma}(s_{o}, \tau_{\sigma, x}) = F^{\sigma}(x) \coloneqq R(x)
\end{equation}
for the only trajectory $\tau_{\sigma, x}$ starting at $s_{o}$ and finishing at $x \in \mathcal{X}$ with positive probability under $p_{F}^{\sigma}$ (as explained earlier, $p_{B}^{\tau}(s, \cdot)$ is either $1$ or $0$). 
Also, it should be clear that  
\begin{equation*}
    \sum_{x \in \mathcal{X}} p_{F}^{\sigma}(s_{o}, \tau_{\sigma,x}) = \sum_{\tau \in s_{o} \rightsquigarrow \mathcal{X}} p_{F}^{\sigma}(s_{o}, \tau) = 1, 
\end{equation*}
as there is only one trajectory from $s_{o}$ to each $x$ and, separating the sum according to the terminal state $x \in \mathcal{X}$ of each $\tau$, this is exactly $\sum_{x \in \mathcal{X}} p_{\top}(x)$ for the $p_{\top}$ introduced in \Cref{eq:marginal}. 
Then, when we sum \Cref{eq:tbindb} over $x \in \mathcal{X}$, \looseness=-1 
\begin{equation*}
    F^{\sigma}(s_{o}) = \sum_{x \in \mathcal{X}} R(x) \coloneqq Z. 
\end{equation*}
This shows $F^{\sigma}(s_{o})$ does not depend on $\sigma$. 
Similarly, to show that \Cref{eq:dbmarginals} is satisfied, we proceed by induction in $i$. 
For this, let $\mathrm{T}(\sigma, s^{(i)}) = \{ x \in \mathcal{X} \colon x_{\sigma([i])} = s^{(i)}_{\sigma([i])} \}$. 
Then, for $i = d$, $s^{(d)} \in \mathcal{X}$, $\mathrm{T}(\sigma, s^{(d)}) = \{s^{(d)}\}$ and $F^{\sigma}(s^{(d)}) = R(x)$ by definition.
Assume, for $i < d$, that $F(s^{(i + 1)}) = \sum_{x \in \mathrm{T}(\sigma, s^{(i + 1)})}  R(x)$ for each $(i + 1)$-sized object $s^{(i + 1)}$. 
Then, summing over the support of $p_{F}^{\sigma}(s_{i}, \cdot)$ in the DB condition in \Cref{eq:dbsigma}, we observe that 
\begin{equation*}
    F^{\sigma}(s^{(i)}) = \sum_{k \in [K]} F^{\sigma}(s^{(i)} \cup \{(k, \sigma(i + 1))\}).  
\end{equation*}
Clearly, $s^{(i), k} = s^{(i)} \cup \{(k, \sigma(i + 1))\}$ corresponds to an $(i + 1)$-sized sequence. 
Also, the sets $\mathrm{T}(\sigma, s^{(i), k})$ are disjoint for $k \in [K]$ as, if $x \in \mathrm{T}(\sigma, s^{(i), k}) \cap \mathrm{T}(\sigma, s^{(i), k'})$, then $x_{\sigma(i + 1)} = k$ and $x_{\sigma{(i + 1)}} = k'$. 
Additionally, $\mathrm{T}(\sigma, s^{(i)}) = \bigcup_{k \in [K]} \mathrm{T}(\sigma, s^{(i), k})$ (see \Cref{fig:pcgflownets}).
By induction, 
\begin{equation*}
    \begin{aligned} 
        F^{\sigma}(s^{(i)}) &= \sum_{k \in [K]} F^{\sigma}(s^{(i), k}) \\ 
        &= \sum_{k \in [K]} \sum_{x \in \mathrm{T}(\sigma, s^{(i), k})} R(x) = \sum_{x \in \mathrm{T}(\sigma, s^{(i)})} R(x). 
    \end{aligned} 
\end{equation*}
In particular, 
\begin{equation*}
    \frac{F^{\sigma}(s^{(i)})}{F^{\sigma}(s_{o})} = \sum_{x \in \mathrm{T}(\sigma, s^{(i)})} \frac{R(x)}{Z},  
\end{equation*}
which is exactly the marginal distribution of $s^{(i)}$ under $R$. 

\subsection{Proof of Corollary~\ref{col:marginals}}

This follows from \Cref{prop:marginals}. 
In fact, notice that $\mathrm{T}(\sigma_{1}, s^{1, (i)}) = \mathrm{T}(\sigma_{2}, s^{2, (i)})$, as if $x \in \mathrm{T}(\sigma_{1}, s^{1, (i)})$, then $x_{\sigma([i])} = s^{1, (i)}_{\sigma([i])} = s^{2, (i)}_{\sigma([i])}$, and so $x \in \mathrm{T}(\sigma_{2}, s^{2, (i)})$, and vice-versa. 
As a consequence, \looseness=-1 
\begin{equation*} 
    \begin{aligned} 
        F^{\sigma_{1}}(s^{1, (i)}) &= \sum_{x \in \mathrm{T}(\sigma_{1}, s^{1, (i)})} R(x) \\ &= \sum_{x \in \mathrm{T}(\sigma_{2}, s^{2, (i)})} R(x) = F^{\sigma_{2}}(s^{2, (i)}).
    \end{aligned} 
\end{equation*} 

\subsection{Proof of Corollary~\ref{col:conditionals}}

This follows from \Cref{col:marginals}.
As both $p_{F}^{\sigma_{1}}$ and $p_{F}^{\sigma_{2}}$ satisfy the DB condition, for $j \in \{1, 2\}$, 
\begin{equation*}
    p_{F}^{\sigma_{j}}(s^{j, (i)}, s^{j, (i + 1)}) = \frac{F^{\sigma_{j}}(s^{j, (i + 1)})}{F^{\sigma_{j}}(s^{j, (i)})}, 
\end{equation*}
with $s^{j, (i + 1)} = s^{j, (i)} \cup \{(k, \sigma_{j}(i + 1))\}$ for some $k \in [K]$, and zero otherwise. 
As $\sigma_{1}(\le i + 1) = \sigma_{2}(\le i + 1)$, \Cref{col:marginals} ensures that $F^{\sigma_{1}}(s^{1, (i)}) = F^{\sigma_{2}}(s^{2, (i)})$ and $F^{\sigma_{1}}(s^{1, (i + 1)}) = F^{\sigma_{2}}(s^{2, (i + 1)})$, and then 
\begin{equation*}
    p_{F}^{\sigma_{1}}(s^{1, (i)}, \cdot) = p_{F}^{\sigma_{2}}(s^{2, (i)}, \cdot). 
\end{equation*}

\subsection{Proof of Proposition~\ref{prop:equivalences}}

This proposition follows directly from \Cref{col:marginals,col:conditionals}, and the notational mapping discussed in \Cref{sec:pcgflownets}.
That said, we provide a comprehensive demonstration below. 

$(\implies)$ 
Let $(p_{\theta}, p_{\phi})$ be a consistent MaM satisfying $p_{\theta}(x) = \pi(x)$ for all $x$. 
Then, define a PC-GFlowNet $(p_{F}^{\sigma}, F^{\sigma})$ as follows.
For each $\mathcal{J} \subseteq \{1, \dots, d\}$, let $\sigma_{\mathcal{J}}$ be a permutation for which the first $|\mathcal{J}|$ elements are $\mathcal{J}$. 
For $i \in \{1, \dots, d\} \setminus \mathcal{J}$, let $\sigma_{\mathcal{J}}^{i}$ be a permutation for which the first $|\mathcal{J}|$ elements are $\mathcal{J}$, and the $(|\mathcal{J}| + 1)$-th element is $i$.

In this context, let $F^{\sigma_{\mathcal{J}}}(x_{\mathcal{J}}) = p_{\theta}(x_{\mathcal{J}})$.
Also, let $\mathcal{I} = \mathcal{J} \cup \{i\}$ and $p_{F}^{\sigma_{\mathcal{J}}^{i}}(x_{\mathcal{J}}, x_{\mathcal{I}}) = p_{\phi}(x_{\mathcal{I}} | x_{\mathcal{J}})$, with $x_{\mathcal{I}}$ and $x_{\mathcal{J}}$ as in Equation (11). 

By Corollaries 3.1 and 3.2, the above construction does not depend on the choice of permutation $\sigma$. 
Consequently, $(p_{F}^{\sigma}, F^{\sigma})$ is equivalent to the MaM $(p_{\theta}, p_{\phi})$.

$(\impliedby)$ 
Conversely, let $(p_{F}^{\sigma}, F^{\sigma})$ be a PC-GFlowNet abiding by the DB condition. 
Let $\sigma_{\mathcal{J}}^{i}$, $\sigma_{\mathcal{J}}^{i}$, and $\mathcal{I}$ be defined as above, and let 
$$
	p_{\theta}(x_{\mathcal{J}}) = F^{\sigma_{\mathcal{J}}(x_{\mathcal{J}}) \text{ and } p_{\phi}(x_{\mathcal{I}} | \mathcal{J}}) = p_{F}(x_{\mathcal{J}}, x_{\mathcal{I}}). 
$$

By Corollaries 3.1 and 3.2, again, the definition above does not depend on permutation $\sigma_{\mathcal{J}}$ and $\sigma_{\mathcal{J}}^{i}$, as long as they satisfy the property that the first $|\mathcal{J}|$ elements are $\mathcal{J}$, and the $(|\mathcal{J}| + 1)$-th is $i$.  

Consequently, $(p_{\theta}, p_{\phi})$ implements the PC-GFlowNet $(p_{F}^{\sigma}, F^{\sigma})$. 

This shows the equivalence.  

\subsection{Proof of Proposition~\ref{prop:dbcon}}

To see this, let $x$ be $\tau$'s terminal state. 
First, notice that $p_{E}^{\sigma}$ simply denotes a
(exploratory) policy such that $p_{E}(s, \cdot)$ and $p_{F}^{\sigma}(s, \cdot)$ have the same support for each $s \in \mathcal{S}$. 
By the definition of PC GFlowNets, each $\tau$ has size $d$ 
and $s_{i} = x_{\sigma([i])}$ for $i \in \{0, \dots, d\}$. 
For conciseness, define 
\begin{equation*}
    \Delta(x, \sigma, i) \coloneqq \left( \log \frac{F(x_{\sigma([i - 1])}) p_{F}^{\sigma}(x_{\sigma([i])} | x_{\sigma([i - 1])})}{F(x_{\sigma([i])})} \right)^{2}.
\end{equation*}
Then, the LHS of \Cref{eq:dbcon} can be written as 
\begin{equation*}  
    \begin{split} 
        \underset{\tau \sim p_{E}^{\sigma}(s_{o}, \cdot)}{\mathbb{E}} \left[ \sum_{1 \le i \le d} w_{i} \Delta(x, \sigma, i) \right] = \!\!\!\! \underset{x \sim q, i \sim \mathrm{Cat}(\mathbf{w})}{\mathbb{E}} \!\!\!\! 
        \left[ \Delta(x, \sigma, i) \right],
    \end{split} 
\end{equation*}
as the marginal of $p_{E}^{\sigma}$ matches $q$.
This is exactly the inner expectation of \Cref{eq:dbcon}'s RHS.

\subsection{Proof of Proposition~\ref{prop:ergodic}}

To ensure that $\{x_{t}\}$ is ergodic, we show that (i) it is irreducible (i.e., every state is reachable from every other state), (ii) aperiodic (i.e., the chain does not return to 
$x_{t}$ at regular intervals), and (iii) stationary with respect to $\pi$ \citep{meyn2009markov}. 
\looseness=-1 

By definition of both $\mathcal{X}$ and $\mathcal{S}$, and since neither $p_{F}$ or $p_{B}$ are degenerate (i.e., they do not assign zero probability to valid transition), the chain $\{x_{t}\}$ is irreducible. 
Additionally, for any $t$, we can return to $x_{t}$ with positive probability after any number of steps.
Hence, the chain is aperiodic. 

To see that the chain is stationary with respect to $\pi$, we first recall that the detailed balance implies 
\begin{equation*}
    F(s) p_{F}(s, x) = R(x) p_{B}(x, s)
\end{equation*}
for any $x \in \mathcal{X}$ and $s \in \mathcal{S}$.
Denote by $\kappa \colon \mathcal{X} \times \mathcal{X} \rightarrow \mathbb{R}_{+}$ the transition kernel of $\{x_{t}\}$. 
Our objective is to show that 
\begin{equation*}
    R(x) \kappa(x, x') = R(x') \kappa(x', x), 
\end{equation*}
which implies $\{x_{t}\}$ is stationary with respect to $\pi(x) \propto R(x)$. 
For this, we notice that 
\begin{equation*}
    \kappa(x, x') = \sum_{s \in \mathcal{S}} p_{B}(x, s) p_{F}(s, x'). 
\end{equation*}
Hence,
\begin{equation*}
    \begin{aligned} 
        R(x) \kappa(x, x') &= \sum_{s \in \mathcal{S}} \textcolor{teal}{R(x) p_{B}(x, s)} p_{F}(s, x') \\ 
        &= \sum_{s \in \mathcal{S}} \textcolor{teal}{F(s) p_{F}(s, x)} p_{F}(s, x') \\ 
        &= \sum_{s \in \mathcal{S}} \textcolor{blue}{F(s)} p_{F}(s, x) \textcolor{blue}{p_{F}(s, x')} \\ 
        &= \sum_{s \in \mathcal{S}} \textcolor{blue}{R(x') p_{B}(x', s)} p_{F}(s, x) \\
        &= R(x') \kappa(x', x); 
    \end{aligned} 
\end{equation*}
we highlight in \textcolor{teal}{teal} and in \textcolor{blue}{blue} the terms for which we apply the DB condition. This shows $\{x_{t}\}$ is stationary with respect to $\pi$. 
Taken together, our results ensure $\{x_{t}\}$ is ergodic with respect to $\pi$. \looseness=-1 



\section{Related works} \label{sec:relatedworks}  

GFlowNets \citep{bengio2021, theory, foundations} are a topic of major interest in the probabilistic modelling literature, providing a clear framework for reasoning about complex hierarchical variational approximations of discrete stochastic models \citep{malkin2023gflownets, choi2025reinforcedsequentialmontecarlo}. 
The central challenge hampering GFlowNets' broader applicability, in our opinion, is that they are often notoriously difficult to train.  
A prominent research direction for mitigating this problem, in fact, has been the development of effective learning objectives \citep{madan2022learninggf, malkin2022trajectory, hu2024squarederrorexploringloss, pan2024pretraining, lingtrajectory} that accelerate training convergence.
As learning efficiency is characterized by not only the objective function, but also by which samples are observed throughout training, another significant recent line of research has explored meta-heuristic approaches for enhanced state space exploration
\citep{lau2023dgfndoublegenerativeflow, kim2025adaptive, madan2025towards, malek2025lossguidedauxiliaryagentsovercoming, dallantonia2026avoidknowdivergenttrajectory}. 
In particular, our method operationally resembles \citet{kim2024localsearchgflownets, hu2023gflownetemlearningcompositionallatent}, both of which introduce a sampling technique based on repeated applications of forward and backward policies; however, only P-GFlowNets asymptotically reduces the number of forward passes per gradient step as a function of the underlying MDP's horizon. 
Nonetheless, despite significantly enhancing GFlowNet's sample efficiency, these approaches often incur in a significant computational cost due to expensive exploration strategies requiring numerous policy evaluations per sample. 
In fact, our evaluation of SA-GFlowNets, Adaptive Teachers GFlowNets, and ACE \citet{madan2025towards, kim2025adaptive, dallantonia2026avoidknowdivergenttrajectory}, alongside Local Search GFlowNets \cite{kim2023local}, suggested an increase of up to $4 \times$ in the per-sample processing time when compared against \citet{madan2022learninggf,malkin2022trajectory}'s traditional algorithms, which remain standard in the GFlowNet literature, e.g., \citep{hu2023amortizing, venkatraman2024amortizingintractableinferencediffusion, zhou2023phylogfn}. 
As in MaMs \citep{liu2023mam}, our work addresses a fundamentally different problem: contrarily to prior approaches, we assume reward querying is cheap, and exploration cost is dominated by policy evaluation. 
This is the dominant setting for Bayesian inference over complex models in large state spaces, e.g., \citet{deleu2022bayesian}, wherein the policy $p_{F}$ is frequently parameterized with inference-intensive models such as a transformer \cite{vaswani2023attentionneed}. 
With this in mind, as discussed in \Cref{sec:discussions}, we believe the ideal algorithm would adaptively provision the appropriate amount of computation for each batch of samples. 
How such an approach would have to be implemented, however, remains open. 
\looseness=-1 


\section{Experimental details} 

Our models were implemented in JAX \citep{jax2018github}. All our experiments were run in a 
Apple MacBook Pro, Apple M4 (10-core: 4P + 6E), 16 GB unified memory, macOS 26.2 (Tahoe).
In \Cref{sec:experiments}, the GFlowNet was trained by minimizing the TB loss \cite{malkin2022trajectory}.
We used Muon optimizer \cite{jordan2024muon} for minimizing the learning objectives for both GFlowNets and P-GFlowNets. 
As in \cite{madan2022learninggf}, we used a learning rate of $10^{-3}$ for $p_{F}$ and of $10^{-2}$ for $F$ and $Z$; $p_{B}$ was fixed as an uniform policy. 
For each experiment, we implemented a 2-layer MLP with 256 hidden units for parameterizing the policy network, and fixed $B = 64$ for the batch size. We trained each model with a fixed time budget shown in \Cref{fig:setsvarsel,fig:AAA,fig:isingbits}.
In particular, we also set $\nu = \eta = 10^{-1}$ for 
the task of Bayesian variable selection and, for the bit sequence generation, followed \citet{malkin2022trajectory}'s approach for constructing the set of modes $\mathcal{M}$. 
All plots show the average across 3 independent runs; error bars represent one standard deviation from the average. 
We periodically evaluated the $\hat{R}$ asynchronously based on the $512$ most recently observed states, following the implementation in Stan \citep{carpenter2017stan}, rejuvenating the chain when the computed $\hat{R}$ exceeded $1.1$. 

\section{Additional Experiments} 

\begin{wrapfigure}[10]{l}{.4\linewidth} 
    \includegraphics[width=\linewidth]{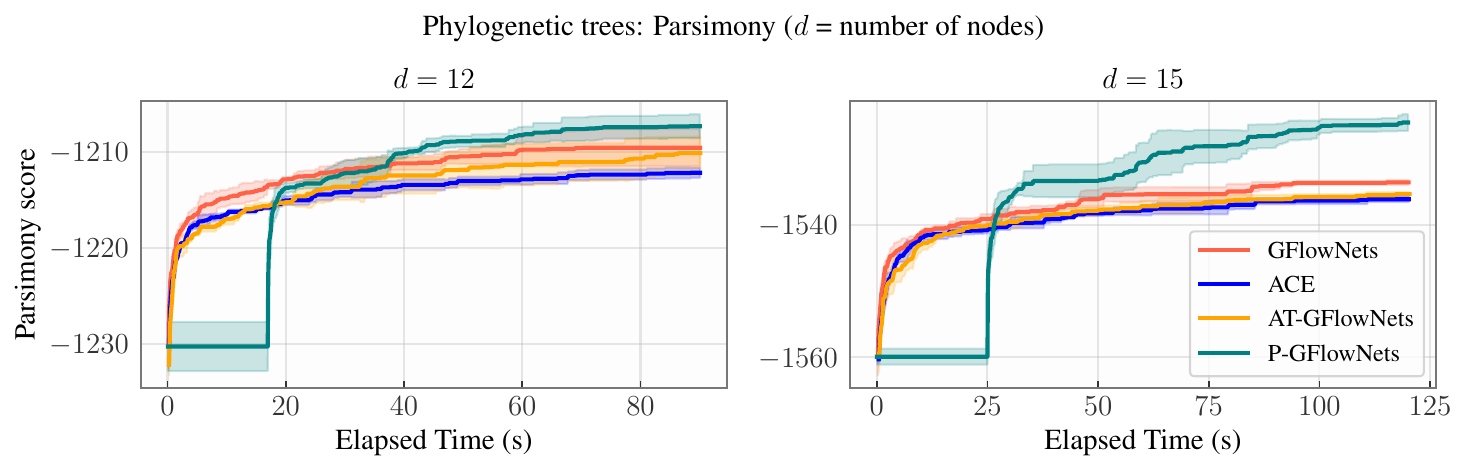}
    \caption{\textbf{P-GFlowNets finds more parsimonious} phylogenetic trees than ACE, AT GFlowNets and TB GFlowNets.}
    \label{fig:treesa} 
\end{wrapfigure}
To further evaluate P-GFlowNets, we compare it against the recently proposed Adaptive Teachers (AT) \cite{kim2025adaptive} and the Adaptive Complementary Exploration (ACE) \cite{dallantonia2026avoidknowdivergenttrajectory} training algorithms. 
We also confirm P-GFlowNets' effectiveness in the phylogenetic inference task \cite{zhou2024phylogfn}. 

\begin{figure*}
    \centering 
    \begin{subfigure}{.25\linewidth}
        \centering
        \includegraphics[width=\linewidth]{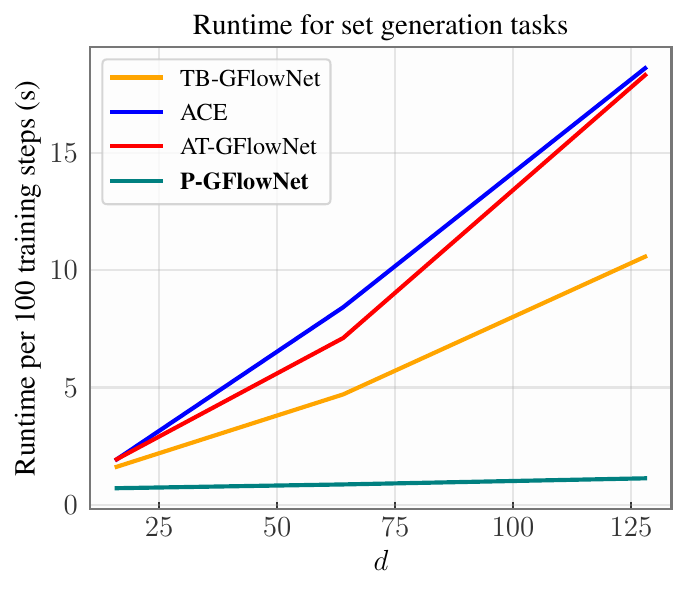}
    \end{subfigure}
    \hspace{58pt} 
    \begin{subfigure}{.25\linewidth}
        \centering 
        \includegraphics[width=\linewidth]{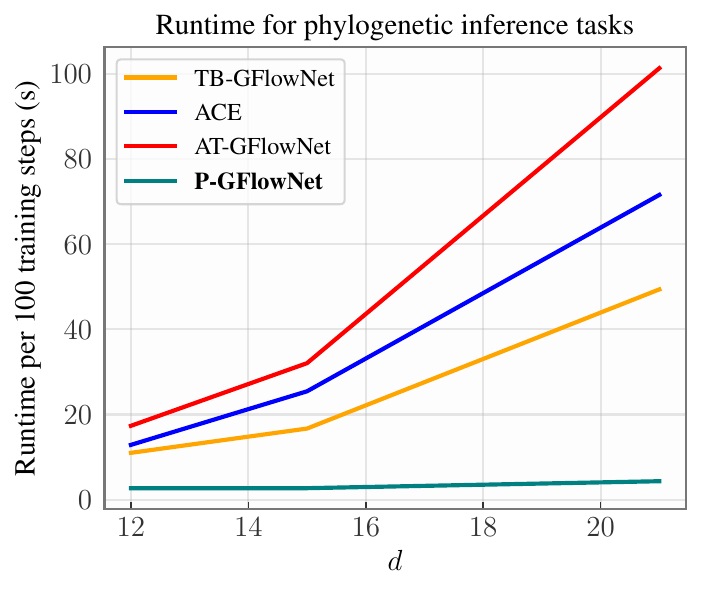}
    \end{subfigure}
    \caption{\textbf{Per-training step runtime for ACE, AT, TB, and P-GFlowNets (ours)}. By circumventing complete trajectory sampling during training, P-GFlowNets reduce per-step computation cost by several orders of magnitude.}
    \label{fig:setsa} 
\end{figure*}

\begin{figure*}
    \begin{subfigure}{.49\linewidth}
        \includegraphics[width=\linewidth]{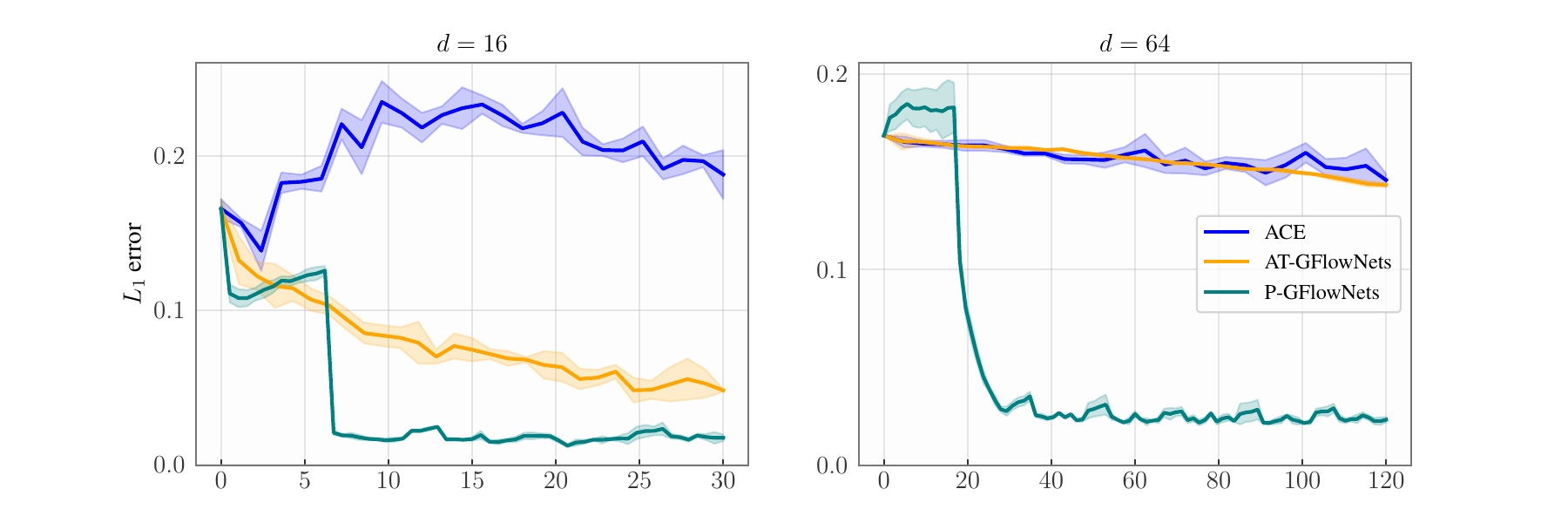}
        \caption{$L_{1}$ error for set generation.}
    \end{subfigure}
    \begin{subfigure}{.49\linewidth}
        \includegraphics[width=\linewidth]{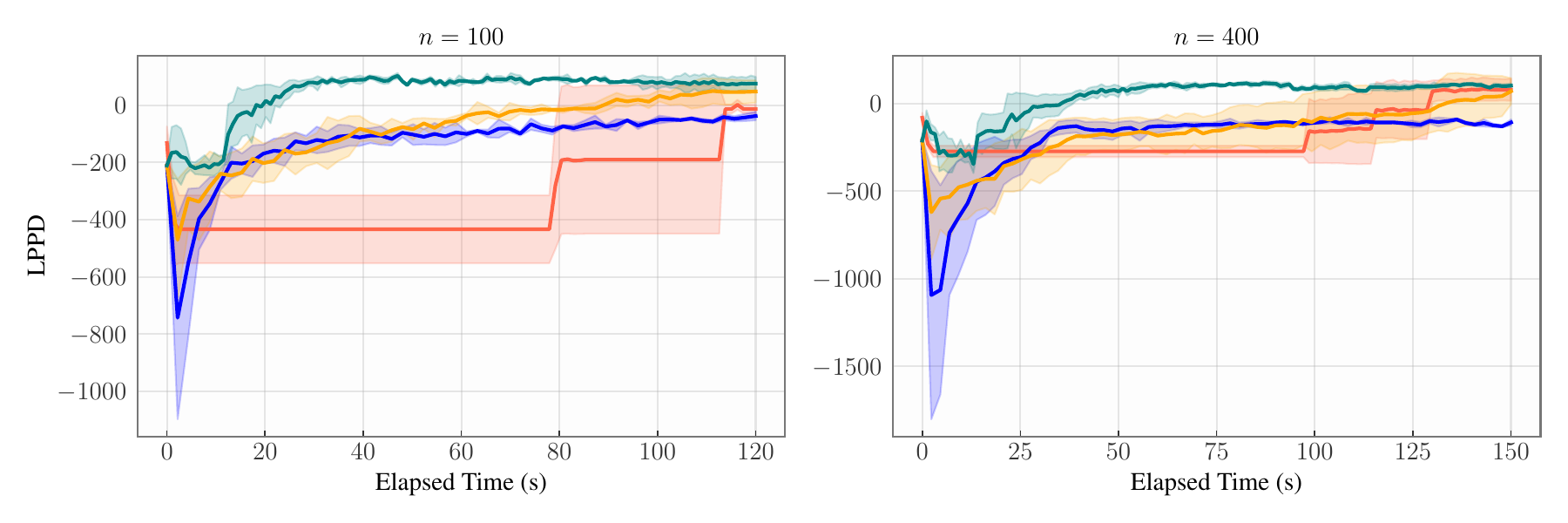}
        \caption{LPPD for variable selection.} 
    \end{subfigure}
    \caption{
        \textbf{P-GFlowNets converge faster than AT, ACE, and TB GFlowNets} in the set generation and variable selection tasks, achieving more accurate marginals (a) and larger log-predictive density (b), respectively, within a shorter time span.
    }
    \label{fig:setsaa} 
\end{figure*}

\paragraph{Comparison against AT and ACE.}
In contrast to P-GFlowNets, whose focus lies on reducing the per-step computational cost for both training and inference, AT and ACE aim at improving a GFlowNet's sample efficiency by training an exploratory model to search for highly informative (e.g., unvisited, high-probability under the target) regions during learning.
In doing so, however, training cost often increases multifold due to the evaluation of both the exploratory and target GFlowNets on both forward and backward trajectories; see \Cref{fig:setsa}.
From this perspective, we found that P-GFlowNets converge significantly faster than AT and ACE with respect to wall-clock time. 
We show illustrate this in \Cref{fig:setsaa} 
for both the set generation and variable selection tasks.
In both cases, we followed the implementations of \cite{dallantonia2026avoidknowdivergenttrajectory} for both AT and ACE GFlowNets. 

The main setting for which these artificial curiosity-inspired strategies would be appropriate, in our opinion, is when the reward function is extremely expensive to evaluate relatively to the policy network \emph{and} this expensiveness cannot be reduced by exploiting the shared structure of adjacent states through caching, as explained next.



\paragraph{Phylogenetic inference.}
To further evaluate P-GFlowNets, we also consider the problem of phylogenetic inference using the parsimony score as the log-reward function \cite{fitch1971toward}. 
We adopt the algorithm suggested by \cite{zhou2024phylogfn}. 
Given a phylogenetic tree $\mathrm{T}$ with leaves $\mathrm{L}$ annotated with $\{\mathrm{A}, \mathrm{T}, \mathrm{C}, \mathrm{G}\}$, let $\mathbf{e}_{n} \in \{1, 0\}^{4}$ be the one-hot encoding of node $n$. 
For $n$ in $\mathrm{L}$, $\mathbf{e}_{n}^{(1)} = 1$ if $n$ is annotated with $A$; $\mathbf{e}_{n}^{(2)} = 1$, if $T$, and etc.
Otherwise, $\mathbf{e}_{n}^{(j)} = 0$. 
Then, by letting $\mathrm{LC}(n)$ and $\mathrm{RC}(n)$ denote the left and right child of $n$, the parsimony score is recursively defined as 
\begin{equation*}
    \begin{aligned} 
        \mathbf{e}_{n} &= 
        \begin{cases} 
            \mathbf{e}_{\mathrm{LC}(n)} \land \mathbf{e}_{\mathrm{RC}(n)}  \text { if } \sum_{i=1}^{4} (\mathbf{e}_{\mathrm{LC}(n)} \land \mathbf{e}_{\mathrm{RC}(n)})^{(i)} \ge 1  \\ 
            \mathbf{e}_{\mathrm{LC}(n)} \lor \mathbf{e}_{\mathrm{RC}(n)}, \text{ otherwise.} 
        \end{cases} \\ 
        \mathrm{ParScore}(n, \mathrm{T}) &= \mathrm{ParScore}(\mathrm{LC}(n), \mathrm{T}) + \mathrm{ParScore}(\mathrm{RC}(n), \mathrm{T}) + [\mathbf{e}_{n} \neq \mathbf{e}_{\mathrm{LC}(n)} \land \mathbf{e}_{\mathrm{RC}(n)}], 
    \end{aligned} 
\end{equation*}
with the boundary condition $\mathrm{ParScore}(n, \mathrm{T}) = 0$ for $n \in \mathrm{L}$ and $\mathbf{u}, \mathbf{v} \mapsto [\mathbf{u} \neq \mathbf{v}]$ being $1$ if $\mathbf{u}$ and $\mathbf{v}$ match element-wise and $0$ otherwise. 
The parsimony score of a tree whose leaves are annotated with a sequence of $\{\mathrm{A}, \mathrm{T}, \mathrm{C}, \mathrm{G}\}$ is the sum of the parsimony score of a tree whose leaves are solely annotated with each element in this sequence, corresponding to the usual bag-of-words assumption in phylogenetic statistical models.   
The intuition is that a tree is as parsimonious (having low parsimony score) as the number of disagreements between a parent and its children. 
We define $R(\mathrm{T}) = \exp\{ - \mathrm{ParScore}(\mathrm{r(\mathrm{T})}, \mathrm{T})\}$ as our reward function, with $\mathrm{r}(\mathrm{T})$ as the root of $\mathrm{T}$. 

Importantly, the modular nature of $\mathrm{ParScore}$ ensures that, in transitioning from $\mathrm{T}$ to $\mathrm{T}'$ using P-GFlowNets' backward-forward kernels, most of the computation required for $\mathrm{ParScore}(\mathrm{T}')$ can be reused from $\mathrm{ParScore}(\mathrm{T})$, reducing the cost of reward querying.
Based on this, we compare P-GFlowNets against AT and ACE GFlowNets on their state space exploration capabilities during training in \Cref{fig:treesa}. 
The initial exploration phase, highlighted as a horizontal line for P-GFlowNets, corresponds to the period before which samples are collected for computing $\hat{R}$, which is used to decide whether the chain should be rejuvenated. 
As in the set generation and variable selection tasks, P-GFlowNets improve upon both AT and ACE given a similar wall-clock time budget.
That said, extending P-GFlowNets to 
mixed, discrete and continuous, spaces---as often required in phylogenetic inference based on stochastic evolution models---remains an interesting venue for future research. 
\looseness=-1 





\end{document}